\documentclass[11pt]{article}

\usepackage[margin=1in]{geometry}
\usepackage{setspace}
\usepackage{amsmath,amssymb,amsfonts}
\usepackage{textcomp}

\usepackage{graphicx}
\usepackage{subcaption}
\usepackage{booktabs}
\usepackage{boldline}
\usepackage[table,xcdraw]{xcolor}

\usepackage[numbers,sort&compress]{natbib}
\usepackage{hyperref}
\hypersetup{
    colorlinks=true,
    linkcolor=blue,
    citecolor=blue,
    urlcolor=blue
}

\usepackage{flushend}
\usepackage{algorithmic}

\title{First Investigation of Deep Learning for Intraoperative Gauze Segmentation in Minimally Invasive Abdominal Surgery}

\author{
    Priya Tomar$^{1,2}$,~
    Maximilian Bro\ss$^{1}$,~
    Philipp Feodorovici$^{3}$,~
    Jan Arensmeyer$^{3}$,\\
    Philipp Leifels$^{2}$,~
    Aditya Parikh$^{1}$,~
    Hanno Matthaei$^{3}$,~
    Christian Bauckhage$^{1,2}$,\\
    Helen Schneider$^{1*}$,~
    Rafet Sifa$^{1,2*}$\\[6pt]
    $^{1}$Fraunhofer IAIS \quad
    $^{2}$University of Bonn \quad
    $^{3}$University Hospital Bonn, Germany\\[4pt]
    \texttt{priya.priya@iais.fraunhofer.de, \{firstname.lastname\}@iais.fraunhofer.de,}\\
    \texttt{ppriya@uni-bonn.de, pleifels@uni-bonn.de, \{firstname.lastname\}@ukbonn.de}\\[4pt]
    {\small $^{*}$Contributed equally}
}

\date{}

\begin{document}

\maketitle

\vspace{-0.5cm}
\begin{center}
\small
\textit{Published in: 2025 IEEE 12th International Conference on Data Science and Advanced Analytics (DSAA)}\\
\textit{DOI:} \href{https://ieeexplore.ieee.org/document/11248028}{10.1109/DSAA65442.2025.11248028}\\[2pt]
\textcopyright~2025 IEEE. Personal use of this material is permitted. Permission from IEEE must be obtained for all other uses.
\end{center}
\vspace{0.3cm}

\begin{abstract}
\noindent
Surgical gauze is an essential part of surgical procedures, primarily used for controlling bleeding and absorbing bodily fluids. The post-surgical retention of gauze can lead to serious complications and necessitate additional surgery for its removal. Despite the clinical significance, research on gauze segmentation using real-world surgical data remains underexplored, owing in part to the scarcity of annotated datasets. In this work, we investigate the use of deep learning methods for gauze segmentation in robot-assisted minimally invasive abdominal surgeries, utilizing an in-house surgical dataset prepared at a university hospital. The training data reflects realistic surgical settings and captures extensive diversity in spatial, morphological, and visual attributes across three different gauze categories. We evaluate several widely used segmentation architectures, including CNN-based, transformer-based, and hybrid architectures, to establish a proof-of-concept for gauze segmentation in a realistic clinical setting. In addition, we investigate the influence of sub-optimally annotated, auto-tracked segmentation masks as a strategy to address data scarcity and improve performance. Our results demonstrate the efficacy of real-world training data in countering the main challenge reported by prior works, the trade-off between blood presence and gauze detection. The incorporation of auto-tracked annotations yields performance enhancements, particularly in generic surgical scenarios. The integration of effective segmentation approaches can benefit robot-guided surgical procedures and various downstream applications by providing precise delineation of foreign objects, thereby enhancing patient safety and surgical outcomes.
\end{abstract}


\section{Introduction}
Operating rooms follow strict standard protocols to avoid unfavorable outcomes and additional complications. Nonetheless, the critical and high-pressure atmosphere in operating rooms can lead to inadvertent errors with serious consequences. One such scenario is the retention of surgical items (RSI) or foreign objects inside the surgical cavity. According to empirical records, RSI occurs most frequently in abdominal surgery, and surgical gauze in particular has the highest retention error rate~\cite{hariharan2013retained}. Commonly used for hemostasis, gauze retention is known as \textit{Gossypiboma} and can result in severe foreign body reactions, including inflammation, sepsis, intestinal blockage, visceral perforations, and potential loss of life~\cite{rajkovic2010unusual,ref_gauze_track,shyam2022gauze}.

Deep learning has made considerable progress in the medical imaging domain, enabling advanced support in automating diagnostic workflows and clinical decision-making~\cite{schneider2022improving, schneider2023one, schneider2023segmentation, schneider2024informed, tomar2025effective}. Current computer vision and deep learning methods for preventing gauze retention primarily rely on convolutional neural network (CNN) based frameworks that exploit textural information using local binary patterns~\cite{ref_LBP} for gauze detection and localization. Although seemingly straightforward, gauze detection in surgical videos is a complex task due to intricate environmental factors such as hematic (blood) presence, stark lighting conditions, dynamic camera movements, smoke, and blood stains on the camera lens. Additionally, the attributes of gauze, its deformable nature, diverse tonalities and texture during surgery, resemblance with illuminated instruments or areas before wound cleaning, and visual blending with surrounding anatomical tissues in case of blood absorption, make detection particularly challenging. Existing works report a trade-off between the presence of blood and gauze detection performance~\cite{ref_gauze_track0,ref_gauze_track,ref_gauze_track_1,ref_gauze_track5}.

In comparison to object detection methods, which mainly provide coarse localizations, segmentation enables advanced support in surgical scene understanding and planning of crucial decisions such as incisions and resections. It enables surgical robots to aid in controlled and safe interaction with anatomical regions, foreign object delineation, and surgical navigation by detecting fine-grained object boundaries~\cite{pinto2023artificial}. Moreover, it helps in targeted interventions by highlighting precise boundaries of pertinent structures while disregarding extraneous information. Additional potential downstream applications of gauze segmentation include real-time decision support enhancement for de-occlusion of foreign objects in augmented reality (AR) based systems~\cite{tanzi2021real,hofman2024first}. AR generates virtual overlays that can occlude critical objects and anatomical regions; de-occlusion in surgical AR can improve surgical safety and help prevent Gossypibomas. Another application involves the extraction and exploitation of gauze features for intra-operative blood loss estimation (BLE) to provide crucial support during operations~\cite{yoon2024automated}. Furthermore, gauze segmentation results can be reduced to typical bounding box detection by extracting enclosing boxes from the respective gauze masks.

To this end, our work investigates effective methods for gauze segmentation in abdominal surgery, with applicability extending to other anatomical and surgical domains. Our contributions are as follows:
\begin{itemize}
    \item Our research presents the first investigation of gauze segmentation on real-world surgical data. The dataset includes three categories of gauze and reflects frequent blood presence.
    
    \item We evaluate several widely used segmentation architectures, including self-attention and transformer-based architectures, which have not been investigated even for gauze detection. In total, we incorporate six architectures spanning pure CNN, vision transformers, and hybrid designs to provide insight into their potential for clinical applicability.
    
    \item We provide a proof-of-concept for using auto-tracked annotations in surgical segmentation, an under-represented research area. Auto-tracked annotations can reduce annotation time and provide additional training data, leading to enhanced performance. Our results confirm their effectiveness, particularly for U-Net variants.
    
    \item Our work demonstrates the significance of utilizing real surgical data in overcoming the challenge reported in existing works regarding poor gauze detection under high blood presence. The models also show robustness to variability in spatial and contextual attributes, as reflected by comparable performance on out-of-domain (OOD) data.
\end{itemize}

\section{Related Work}
The conventional intraoperative method for tracking surgical items entails manual counting before and after surgery by assistants, making it highly prone to inadvertent errors. In case of unbalanced counting post-surgery, gauze detection is conventionally carried out by radiography, where some works propose to complement the gauze with non-deformable radiopaque tags or three-dimensional micro tags to enhance its visibility~\cite{ref_gauze_diag1,ref_gauze_diag2,ref_gauze_diag3}. More advanced approaches suggest equipping gauze with unique radio frequency identification (RFID) tags that are scanned before and after insertion into the surgical cavity~\cite{ref_gauze_det1,ref_gauze_det2,ref_gauze_det3}. Besides causing delays in operative procedures and exposing the patient to radiation, these methods exhibit limited effectiveness due to the involved technical intricacies~\cite{ref_gauze_track}.

Existing works in computer vision primarily focus on exploiting the textural information of gauze for feature extraction through local binary patterns (LBP)~\cite{ref_LBP,ref_gauze_track_1,ref_gauze_track0,ref_gauze_track,yoon2024automated} or optical flow~\cite{pragadeeswari2020optical}. Proposed deep learning methods are limited to CNN architectures, primarily for gauze detection using You Only Look Once (YOLO)~\cite{redmon2016you,ref_gauze_track_1,ref_gauze_track5}. There is extremely limited work on gauze segmentation. S{\'a}nchez-Brizuela et al.~\cite{ref_gauze_track_1} focus on gauze segmentation using U-Net~\cite{ronneberger2015u}, and there are no studies investigating advanced architectures, such as vision transformers, for performance comparisons and applicability. Moreover, the data in~\cite{ref_gauze_track_1} is simulated in a highly controlled environment on animal organs, featuring restricted camera movements, minimal presence of blood, large gauze dimensions, and limited visibility of anatomical regions. Furthermore, existing works on gauze detection report selective applicability due to their suboptimal performance in detecting blood-soaked gauze.

Current surgical segmentation methods focus primarily on instrument segmentation and organ segmentation~\cite{rueckert2024methods,kamtam2025deep}. Research on intraoperative gauze retention is limited to object detection approaches due to insufficient availability of pertinent segmentation datasets. Besides preventing gauze retention, its discrimination from surrounding regions is of paramount importance for downstream applications such as intraoperative blood loss estimation~\cite{yoon2024automated} and de-occlusion~\cite{hofman2024first} for planning crucial operative decisions. This work provides a proof-of-concept exploring the potential performance and applicability of current deep learning models for effective gauze segmentation in a real surgical setting.

\section{Data}

To the best of our knowledge, there exists no open-source dataset prepared from real laparoscopic surgeries for gauze segmentation. The only public dataset containing gauze masks is that of S{\'a}nchez-Brizuela et al.~\cite{gauze_seg_dataset}, which was prepared using a laparoscopic simulator on animal organs, presenting an unrealistic scenario. Therefore, we use an in-house dataset prepared at the University Hospital Bonn (UKB) for training and primary evaluation while utilizing~\cite{gauze_seg_dataset} for additional testing. Henceforth, we refer to the simulated dataset as GauzeSeg$_{Sim}$ and the real-world UKB dataset as GauzeSeg$_{Real}$ (Figure~\ref{fig:dataset_illustration}).

\subsection{GauzeSeg$_{Real}$ Dataset}
The dataset was prepared at the Bonn Surgical Technology Center\footnote{\href{https://boster-bonn.com/}{Bonn Surgical Technology Center (BOSTER)}} in the University Hospital Bonn (UKB) under the supervision of three surgeons. Written informed consent was waived with institutional review board approval (AZ 306/23-EP). Data has been processed in accordance with the Health Data Protection Act of North Rhine-Westphalia (GDSG NW) \S6~(2). Video records of 16 robot-assisted laparoscopic surgeries performed at UKB were selected and trimmed to the resection phase. Video frames were annotated using Supervisely by four research assistants for three types of gauze: conventional gauze, cigar gauze, and swab. The extracted image dimensions are \mbox{1080~$\times$~1920} pixels. Individual samples were annotated by a single annotator; however, to mitigate errors, an annotation atlas was prepared and provided to the annotators as a guide, and the annotation process was strictly monitored by domain experts for quality control.

\begin{figure*}[!hbt]
    \centering
    \includegraphics[width=\textwidth]{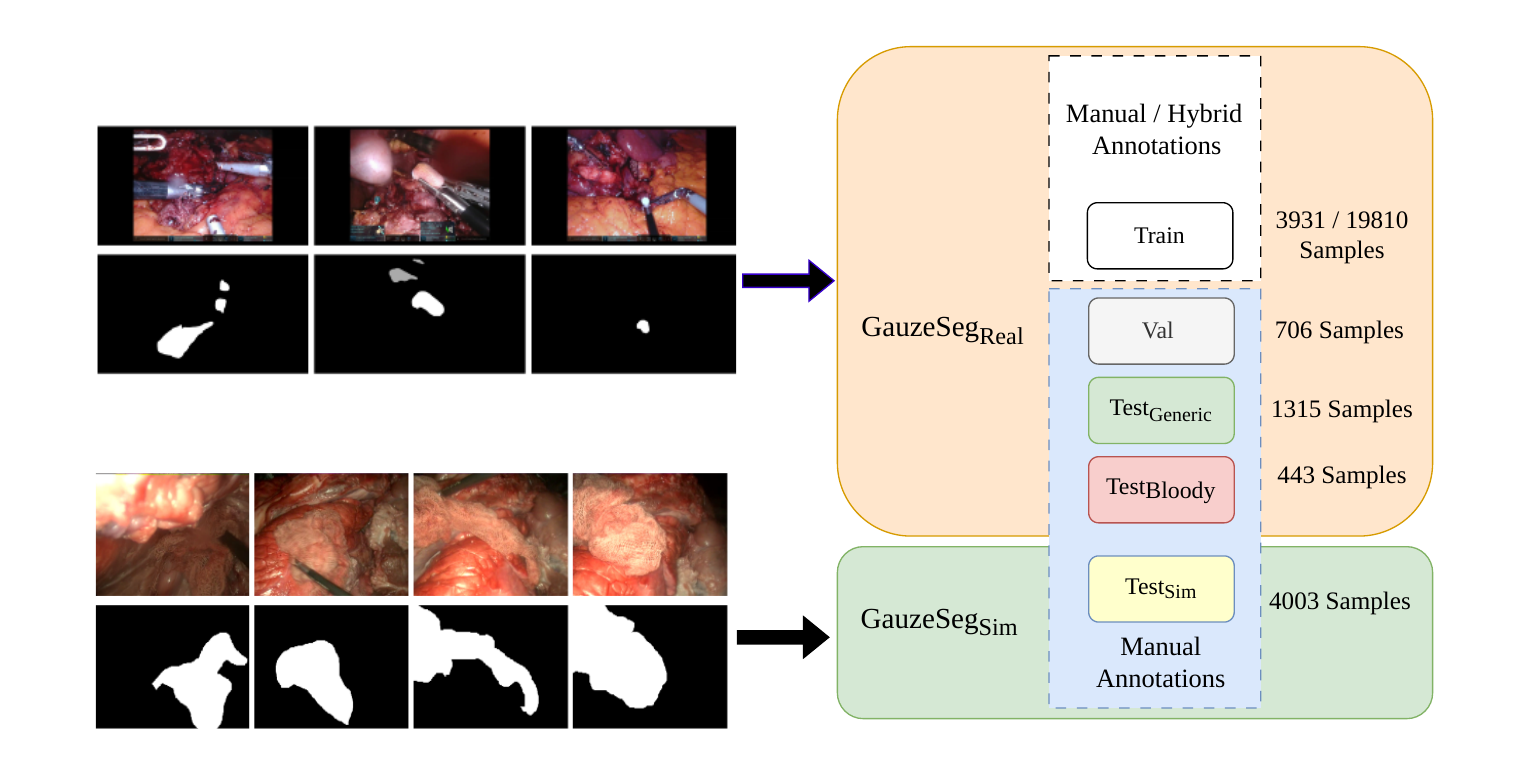}
    \caption{Overview of the GauzeSeg$_{Real}$ (top grid) and GauzeSeg$_{Sim}$ (bottom grid) datasets with respective sample images and annotation masks. The GauzeSeg$_{Real}$ dataset contains real-world surgical frames with manual and hybrid (manual + auto-tracked) annotations, split into training, validation, and test sets by patient ID. Test$_{Generic}$ and Test$_{Bloody}$ represent generalized and high blood presence scenarios, respectively. The GauzeSeg$_{Sim}$ dataset contains simulated surgical frames with manually traced annotations, used entirely as an out-of-distribution test set (Test$_{Sim}$). All evaluation splits contain only manually annotated samples.}
    \label{fig:dataset_illustration}
\end{figure*}

As the data contains images of real robot-assisted laparoscopic surgeries, it represents diverse surgical scenarios and heterogeneity in the type, dimension, location, and visual attributes of gauze, as illustrated in Figure~\ref{fig:dataset_illustration}. It also reflects the frequent presence of multiple gauzes in a single frame, which are often occluded by surgical instruments and anatomical structures.

Supervisely offers an auto-tracking feature that propagates object annotations in videos to subsequent frames based on an initial manual annotation. We leveraged this feature as it speeds up the annotation process and provides additional data for model training. Since the auto-tracked masks represent sub-optimal quality annotations in terms of object contours, they also offer an opportunity to investigate the effect of such semi-automatic annotations on segmentation performance. In total, the dataset comprises 22,265 samples with hybrid annotations and 6,395 manually annotated samples. Figure~\ref{fig:cls_distribution} shows the per-patient distribution of manual and hybrid annotations, illustrating how the auto-tracked masks substantially augment the training data.

We consider two sets for testing: the first set, \mbox{Test$_{Generic}$}, depicts a generalized representation of hematic presence on gauze, from blood-tinged to blood-saturated. The second set, \mbox{Test$_{Bloody}$}, evidently reflects high blood presence in the frames. The videos for both test sets were selected by the surgeons based on the visual attributes of gauze and the presence of blood in the surgical cavity. Refer to Section~\ref{sec:evaluation} for training and evaluation sample counts.

Please note that this study aims to provide a preliminary insight into gauze segmentation on laparoscopic videos in a realistic environment. The dataset will be made publicly available in the future after securing the required regulatory approvals.

\subsection{GauzeSeg$_{Sim}$ Dataset}
The dataset was prepared by simulating diverse operative scenarios on animal organs exported from a slaughterhouse using a laparoscopic simulator. The resultant segmentation masks feature surgical tools with single gauze (clean and stained state) or multiple gauzes in diverse conditions. In total, the dataset contains 4,003 frames with binary masks traced manually, and the image dimensions are \mbox{720$\times$~576} pixels. In comparison to the GauzeSeg$_{Real}$ dataset, where gauze has high diversity in its spatial and visual attributes, gauze in GauzeSeg$_{Sim}$ appears to be the central object in the images and reflects the simulated conditions, as presented in Figure~\ref{fig:dataset_illustration}. Refer to~\cite{gauze_seg_dataset} for additional information concerning this dataset.

\section{Methods and Experiments}

\subsection{Data Pre-processing}
GauzeSeg$_{Real}$ samples contain black regions surrounding the surgical view, as illustrated in Figure~\ref{fig:dataset_illustration}, which we remove before further data pre-processing. During training, we use standard data augmentation approaches such as random cropping, horizontal and vertical flipping, Gaussian blurring, and color jittering, before resizing the inputs to \mbox{512~$\times$~512} dimensions. As the GauzeSeg$_{Sim}$ dataset is already cropped to the surgical view of interest and is utilized only for testing, we do not apply pre-processing beyond input resizing. During our initial experiments, we observed comparatively poor scores for input size 256~$\times$~256, as the decrease in resolution causes a reduction in gauze features. Therefore, all experiments are conducted on input size 512~$\times$~512 pixels.

\subsection{Training Framework}
Due to the absence of prior work on gauze segmentation on real-world surgical data that could serve as a baseline, we investigate several widely used segmentation architectures, including CNN and transformer architectures, to appraise their preliminary performances. We focus on three types of architectures: CNN-based, including \mbox{U-Net}~\cite{ronneberger2015u} and \mbox{DeepLabV3} (DeepLab)~\cite{chen2017rethinking}; transformer-based, including SegFormer~\cite{xie2021segformer} and \mbox{Mask2Former}~\cite{cheng2022masked}; and hybrid architectures, including Attention U-Net (AU-Net)~\cite{oktay2018attention} and TransU-Net~\cite{chen2021transunet}. Although some of these architectures have been investigated for surgical data segmentation, we consider them for gauze segmentation as gauze exhibits greater heterogeneity in its visual and spatial features than other anatomical structures or surgical objects, and the comparison provides a framework covering simplicity, self-attention, transformer architecture, resource constraints, and performance considerations.

In addition to training from scratch for task-specific learning, we investigate the benefits of transfer learning. Besides random weight initializations, we include SegFormer and Mask2Former with pretrained \mbox{ADE20K} weights, and \mbox{DeepLab} with ImageNet-1K weights (referred henceforth as \mbox{SegFormer$_{pt}$}, \mbox{Mask2Former$_{pt}$}, and \mbox{DeepLab$_{pt}$}).
Out of the different available \mbox{SegFormer} models, we incorporate two: \mbox{SegFormer MiT-b1} and \mbox{SegFormer MiT-b3}. For \mbox{DeepLab}, we use the \mbox{ResNet-50} backbone. TransU-Net includes a CNN module and a transformer block. Please refer to the Github repository\footnote{\url{https://github.com/PriyaTomar/GauzeSeg}} for detailed model configurations and reproducibility of experiments. The trainable parameters for U-Net, AU-Net, TransU-Net, and SegFormer-b1 are in a similar range: 13.4M, 13.7M, 13.5M, and 13.7M, respectively. Similarly, the parameters for Mask2Former and SegFormer-b3 are in a similar range at 47.4M and 47.2M. DeepLab has 42.1M trainable parameters.

While nnU-Net~\cite{isensee2021nnu} has demonstrated strong potential in medical image segmentation tasks due to its effective training optimization framework~\cite{isensee2024nnu}, we defer its investigation to future work, as rigorous network and hyperparameter optimization for the included architectures already demands extensive computing resources. This study aims to investigate established and prominently utilized architectures for gauze segmentation to provide preliminary insights into their clinical applicability.

\subsection{Evaluation}\label{sec:evaluation}
Unlike random splitting, we split the data for training and evaluation according to patient IDs (Table~\ref{tab:data_splits}). This approach prevents leakage of training data features to the evaluation splits and reflects a realistic scenario in which the model is applied to a different patient's surgery. For testing, we consider two sets from the \mbox{GauzeSeg$_{Real}$} dataset, selected by the surgeons: the first set, Test$_{Generic}$, represents a generalized operating condition for hematic presence on gauze, from blood-tinged to blood-saturated; the second set, \mbox{Test$_{Bloody}$}, represents high blood presence. The reason for including an additional \mbox{Test$_{Bloody}$} set is to evaluate model performance in high blood presence, given the limitations reported in previous works on gauze detection, where a trade-off is observed between detection performance and blood presence~\cite{ref_gauze_track,ref_gauze_track0,ref_gauze_track_1,ref_gauze_track5}.

GauzeSeg$_{Sim}$ is used entirely as a test set, namely Test$_{Sim}$, for examining segmentation results in comparison with existing work on gauze segmentation. Additionally, it enables validation of the efficacy of our \mbox{GauzeSeg$_{Real}$} dataset in learning suitable features for gauze segmentation on out-of-domain data. Table~\ref{tab:data_splits} presents the sample counts for training, validation, and test sets. The training set with manual and hybrid annotations contains 3,931 and 19,801 samples, respectively, from 6 patient videos. The validation set has 706 samples from two videos. \mbox{Test$_{Generic}$}, \mbox{Test$_{Bloody}$}, and Test$_{Sim}$ contain 1,315, 443, and 4,003 samples, respectively. To ensure reliable and effective evaluation, validation and test sets contain only manually annotated data.

\begin{table}[tb]
\caption{Sample counts and respective patient IDs in training and evaluation sets for \mbox{GauzeSeg$_{Real}$} dataset. Test sets \mbox{Test$_{Generic}$} and \mbox{Test$_{Bloody}$} contain samples from the real-world \mbox{GauzeSeg$_{Real}$} dataset. The third test set Test$_{Sim}$ comprises samples from the simulated \mbox{GauzeSeg$_{Sim}$} setting for additional evaluation on out-of-distribution data. Evaluation splits contain only manually annotated samples.}
\begin{center}
\begin{tabular}{p{2cm}|p{2.3cm}|p{1cm}|p{1cm}}
 \bfseries Split & \bfseries Patient ID &  \multicolumn{2}{c}{\bfseries Annotation}   \\
 \toprule
  & & Manual  & Hybrid \\
    \cmidrule{3-4}
     Training & 1, 5, 6, 9, 14, 16 & 3931  & 19801 \\
     Validation & 10, 22   &  706 & - \\
     Test$_{Generic}$ & 3, 11, 17, 23  & 1315  & - \\
     Test$_{Bloody}$ & 4, 18, 19, 21  &  443 &  -\\
     Test$_{Sim}$ & -  &  4003 &  -\\
\end{tabular}
\label{tab:data_splits}
\end{center}
\end{table}

As performance metrics, we evaluate the mean values of dice score (Dice), Intersection over Union (IoU), Precision, and Recall. The Dice score is used for the selection of the best model checkpoint on the validation split. Additionally, we use normalized surface distance (NSD)~\cite{nikolov2021clinically,seidlitz2022robust} for analyzing segmentation contours.

\subsection{Experiment Setup}
The indicated parameter values were determined through initial hyperparameter optimization experiments on manual annotations, given resource constraints in conducting an extensive hyperparameter search. To appraise the considered training architectures, we train 5 models per configuration and report the average performance. This also applies to transfer learning approaches, as non-deterministic CUDA operations cause variations in results. Each experiment is initialized with a fixed seed for reproducibility and trained for a maximum of 150 epochs with the \mbox{AdamW} optimizer, weight decay value of 0.1, dice loss function, batch size of 64, and an initial learning rate of $1 \times 10^{-4}$, which is decreased by a factor of 0.1 if no improvement is observed in the validation loss for 10 epochs. All experiments are conducted on two NVIDIA RTX 6000 GPUs with 48\,GB memory using the PyTorch v2.3.1 framework.

As the validation set includes samples from different patient surgeries, we observed high fluctuation in validation scores during initial experiments. Accordingly, we average the validation score with a step size of 5 and use it for early stopping with a patience factor of 25 epochs. Refer to the Github repository\footnote{\url{https://github.com/PriyaTomar/GauzeSeg}} for extensive experiment configurations and reproducibility.

In total, 100 training runs were conducted (10 model configurations $\times$ 2 annotation settings $\times$ 5 seeds). The total computational cost amounted to approximately 1,400 GPU-hours on two NVIDIA RTX 6000 GPUs (48\,GB each), corresponding to an estimated energy consumption of approximately 410\,kWh based on a thermal design power of 300\,W per GPU.

\section{Results}
\subsection{Manual Annotations}

\begin{table}[!tb]
\caption{Results on manual annotations for different architectures. \textbf{Bold} indicates the highest scores in respective test sets. The abbreviation $pt$ indicates models initialized with pre-trained weights. Mask2Former and SegFormer experience underfitting when trained from scratch, but we include the results for comparison.}
\label{tab:baseline}
\centering
\begin{tabular}{llcc}
\textbf{Model} & \textbf{Test Set} & \textbf{Dice} & \textbf{IoU} \\ \toprule
\rowcolor{gray!20}U-Net & Test$_{Generic}$ & 0.43 $\pm$ 0.01 & 0.37 $\pm$ 0.01 \\
\rowcolor{gray!20}      & Test$_{Bloody}$  & 0.63 $\pm$ 0.03 & 0.49 $\pm$ 0.04 \\
\rowcolor{gray!20}      & Test$_{Sim}$     & 0.69 $\pm$ 0.03 & 0.56 $\pm$ 0.04 \\ \midrule
\rowcolor{gray!20}DeepLab & Test$_{Generic}$ & 0.70 $\pm$ 0.06 & 0.63 $\pm$ 0.06 \\
\rowcolor{gray!20}       & Test$_{Bloody}$  & 0.83 $\pm$ 0.02 & 0.72 $\pm$ 0.02 \\
\rowcolor{gray!20}       & Test$_{Sim}$     & 0.74 $\pm$ 0.04 & 0.63 $\pm$ 0.04 \\ \midrule
\rowcolor{gray!20}DeepLab$_{pt}$ & Test$_{Generic}$ & 0.70 $\pm$ 0.01 & 0.62 $\pm$ 0.01 \\
\rowcolor{gray!20}               & Test$_{Bloody}$  & 0.80 $\pm$ 0.03 & 0.69 $\pm$ 0.03 \\
\rowcolor{gray!20}               & Test$_{Sim}$     & 0.75 $\pm$ 0.03 & 0.64 $\pm$ 0.04 \\ \specialrule{1pt}{4pt}{4pt}
\rowcolor{brown!10}AU-Net & Test$_{Generic}$ & 0.44 $\pm$ 0.02 & 0.37 $\pm$ 0.02 \\
\rowcolor{brown!10}       & Test$_{Bloody}$  & 0.66 $\pm$ 0.02 & 0.51 $\pm$ 0.02 \\
\rowcolor{brown!10}       & Test$_{Sim}$     & 0.66 $\pm$ 0.03 & 0.52 $\pm$ 0.04 \\ \midrule
\rowcolor{brown!10}TransU-Net & Test$_{Generic}$ & 0.42 $\pm$ 0.01 & 0.35 $\pm$ 0.01 \\
\rowcolor{brown!10}          & Test$_{Bloody}$  & 0.62 $\pm$ 0.02 & 0.47 $\pm$ 0.02 \\
\rowcolor{brown!10}          & Test$_{Sim}$     & 0.58 $\pm$ 0.06 & 0.44 $\pm$ 0.06 \\ \specialrule{1pt}{4pt}{4pt}
Mask2Former & Test$_{Generic}$ & 0.15 $\pm$ 0.05 & 0.09 $\pm$ 0.04 \\
            & Test$_{Bloody}$  & 0.16 $\pm$ 0.03 & 0.09 $\pm$ 0.02 \\
            & Test$_{Sim}$     & 0.44 $\pm$ 0.00 & 0.31 $\pm$ 0.00 \\ \midrule
\rowcolor{red!5}Mask2Former$_{pt}$ & Test$_{Generic}$ & 0.78 $\pm$ 0.02 & 0.68 $\pm$ 0.02 \\
\rowcolor{red!5}                 & Test$_{Bloody}$  & 0.77 $\pm$ 0.03 & 0.63 $\pm$ 0.03 \\
\rowcolor{red!5}                 & Test$_{Sim}$     & 0.82 $\pm$ 0.01 & 0.71 $\pm$ 0.02 \\ \midrule
SegFormer-b1 & Test$_{Generic}$ & 0.22 $\pm$ 0.03 & 0.15 $\pm$ 0.02 \\
             & Test$_{Bloody}$  & 0.02 $\pm$ 0.01 & 0.01 $\pm$ 0.01 \\
             & Test$_{Sim}$     & 0.15 $\pm$ 0.06 & 0.11 $\pm$ 0.05 \\ \midrule
\rowcolor{red!8}SegFormer-b1$_{pt}$ & Test$_{Generic}$ & 0.78 $\pm$ 0.06 & 0.69 $\pm$ 0.06 \\
\rowcolor{red!8}                 & Test$_{Bloody}$  & 0.81 $\pm$ 0.02 & 0.69 $\pm$ 0.03 \\
\rowcolor{red!8}                 & Test$_{Sim}$    & 0.72 $\pm$ 0.05 & 0.62 $\pm$ 0.05 \\ \midrule
\rowcolor{red!8}SegFormer-b3$_{pt}$ & Test$_{Generic}$ & \textbf{0.82 $\pm$ 0.03} & \textbf{0.75 $\pm$ 0.03} \\
\rowcolor{red!8}                 & Test$_{Bloody}$  & \textbf{0.85 $\pm$ 0.01} & \textbf{0.75 $\pm$ 0.02} \\
\rowcolor{red!8}                 & Test$_{Sim}$     & \textbf{0.82 $\pm$ 0.02} & \textbf{0.72 $\pm$ 0.02} \\
\end{tabular}
\end{table}

\begin{figure*}[!ht]
\centering
\begin{subfigure}[t]{0.32\textwidth}
    \centering
    \includegraphics[width=\linewidth]{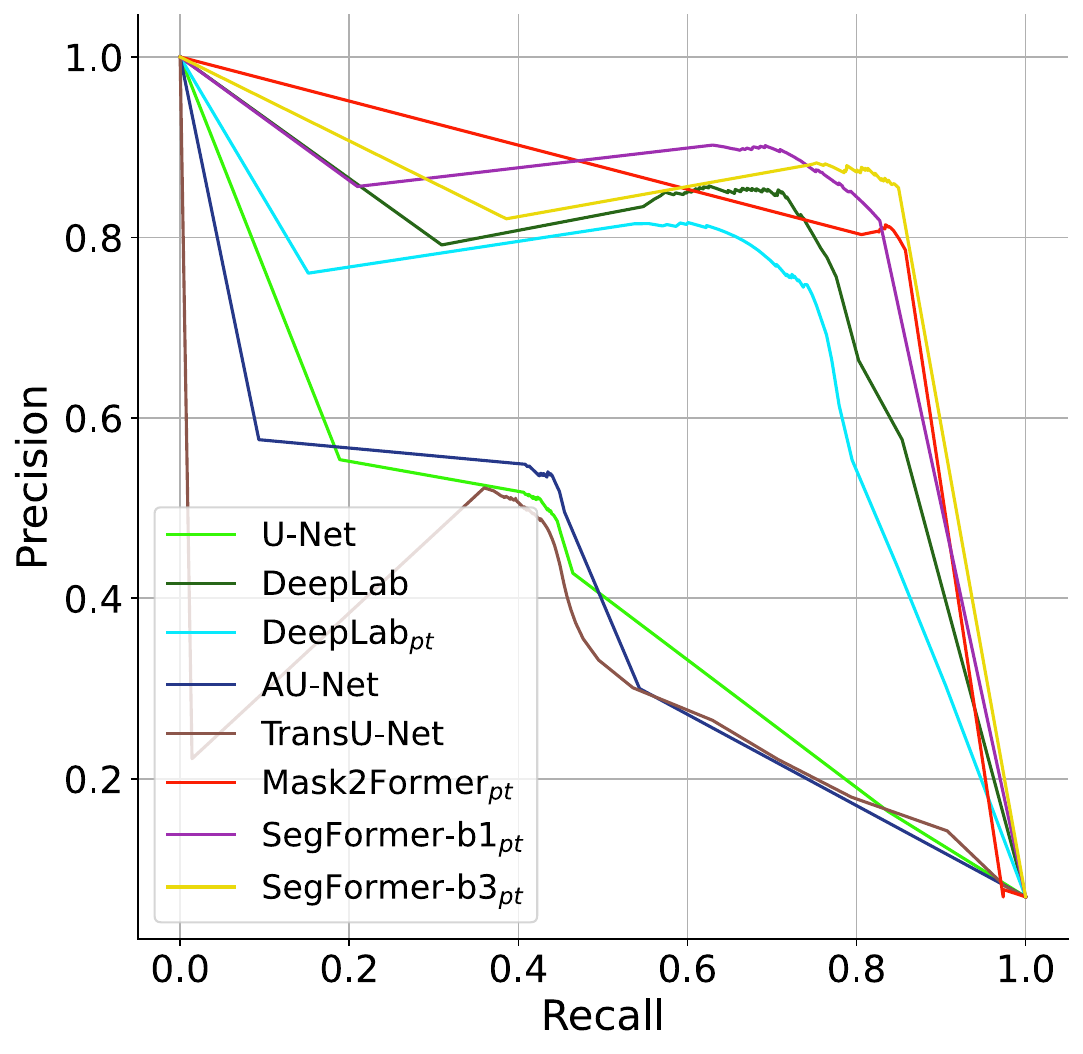}
    \caption{Test$_{Generic}$}
    \label{fig:pr_curve_manual_t1}
\end{subfigure}
\begin{subfigure}[t]{0.32\textwidth}
    \centering
    \includegraphics[width=\linewidth]{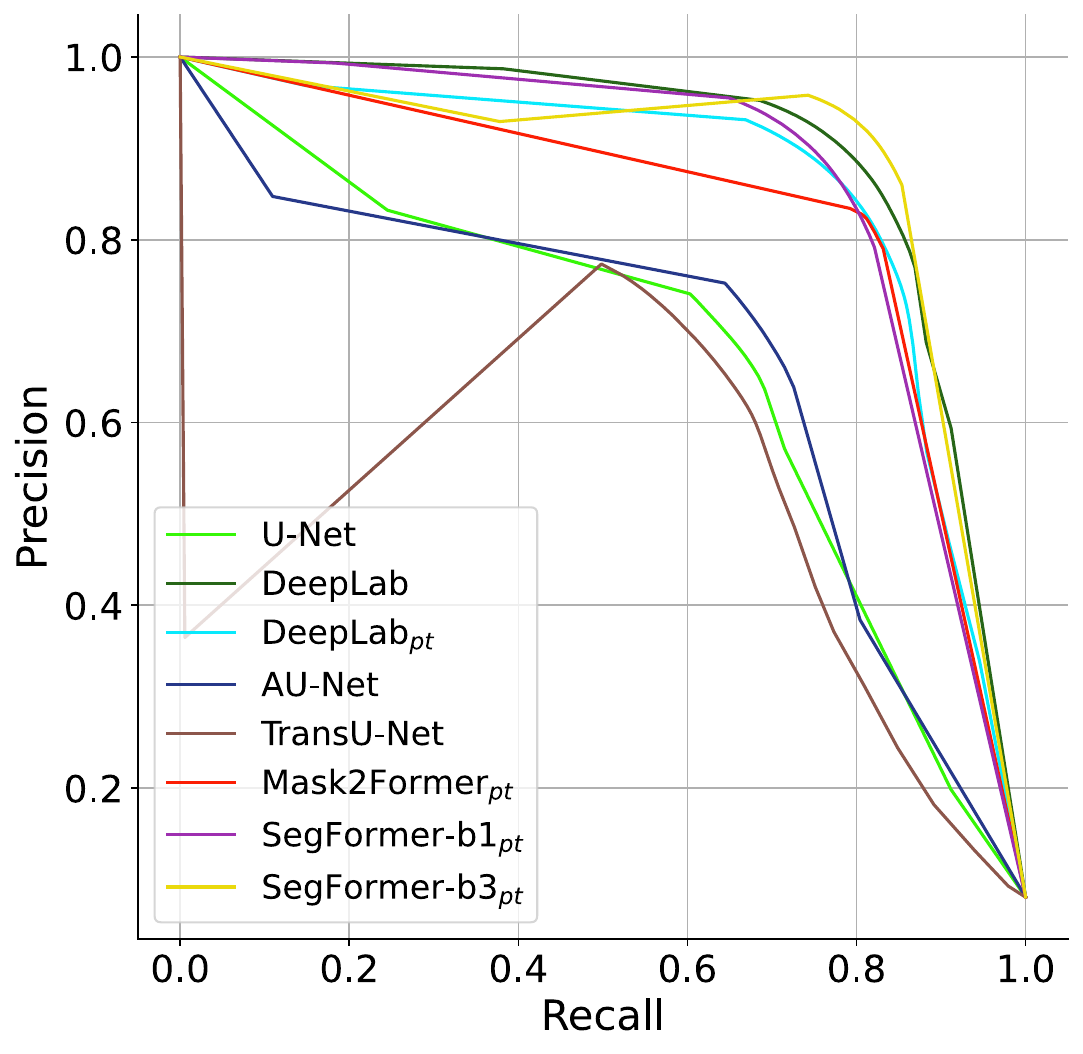}
    \caption{Test$_{Bloody}$}
    \label{fig:pr_curve_manual_t2}
\end{subfigure}
\begin{subfigure}[t]{0.32\textwidth}
    \centering
    \includegraphics[width=\linewidth]{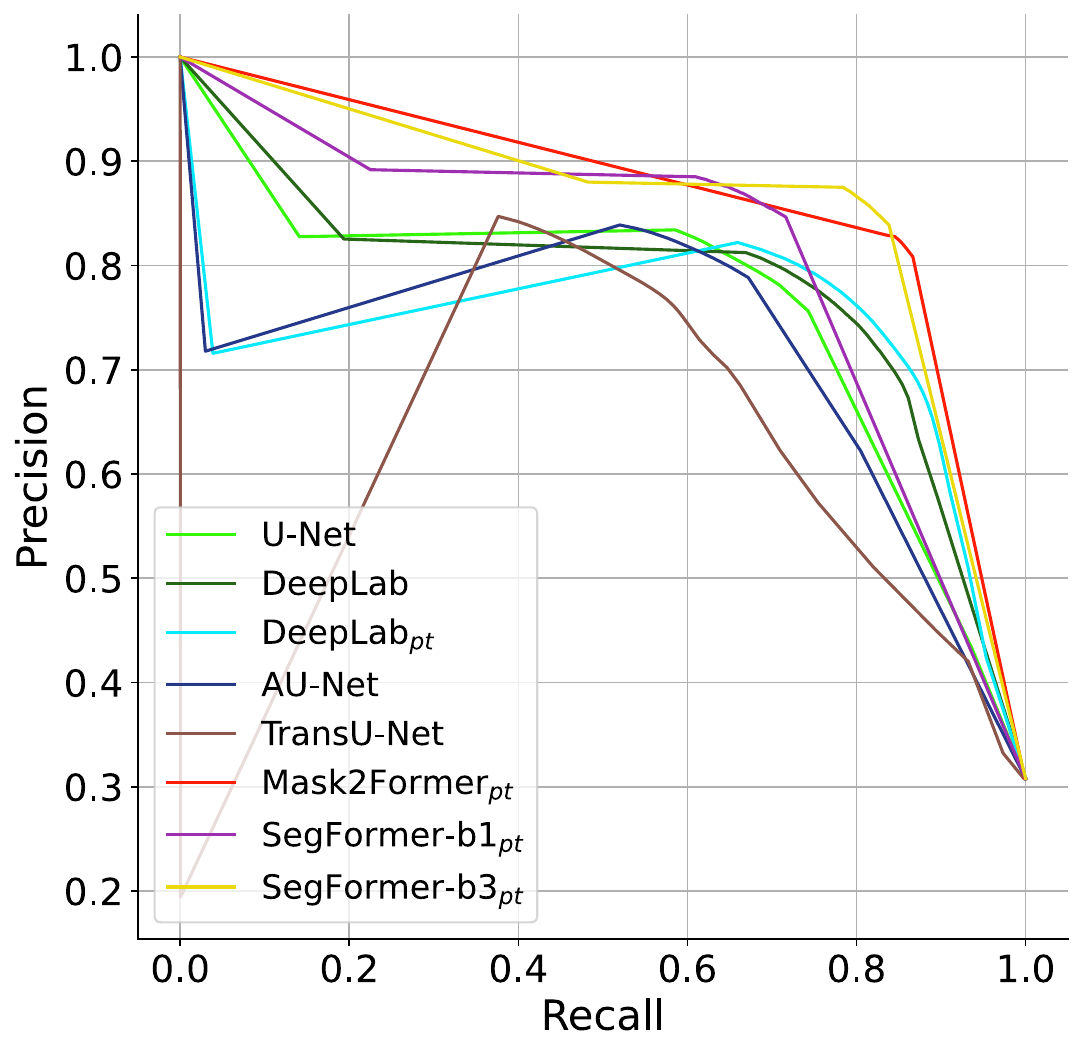}
    \caption{Test$_{Sim}$}
    \label{fig:pr_curve_manual_t3}
\end{subfigure}
\caption{Precision-Recall curves for models trained on manual annotations. Values are averaged over 5 training runs per model. SegFormer-b3$_{pt}$ demonstrates high recall and precision values across all three test sets. Training on real surgical data results in superior performance on the Test$_{Bloody}$ set.}
\label{fig:pr_curve_manual}
\end{figure*}

\begin{figure*}[!ht]
    \centering
    \begin{subfigure}[t]{0.32\textwidth}
        \centering
        \includegraphics[width=\textwidth]{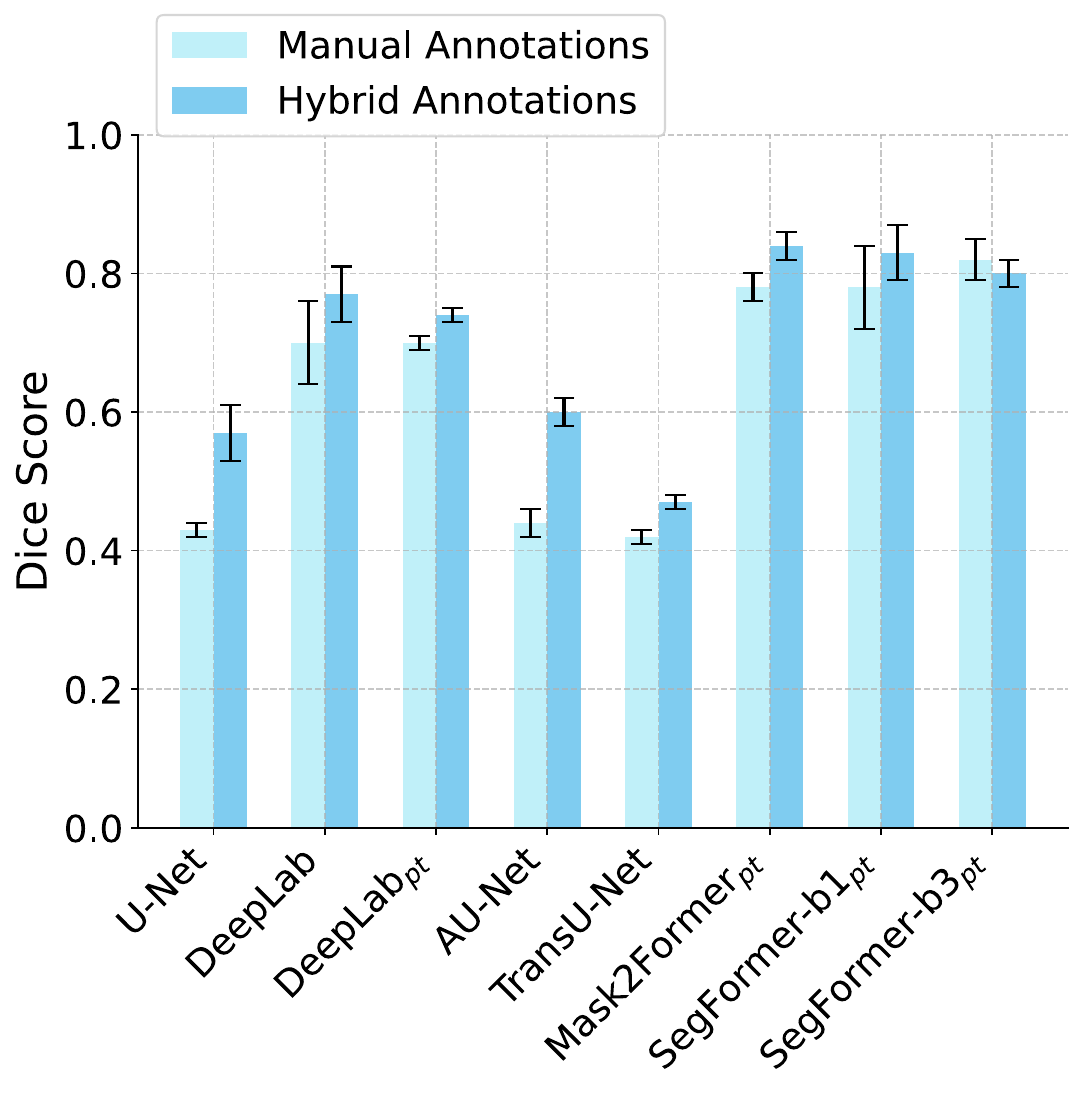}
        \caption{Test$_{Generic}$}
        \label{fig:dice_score_manual_mix_t1}
    \end{subfigure}
     \hfill
    \begin{subfigure}[t]{0.32\textwidth}
        \centering
        \includegraphics[width=\textwidth]{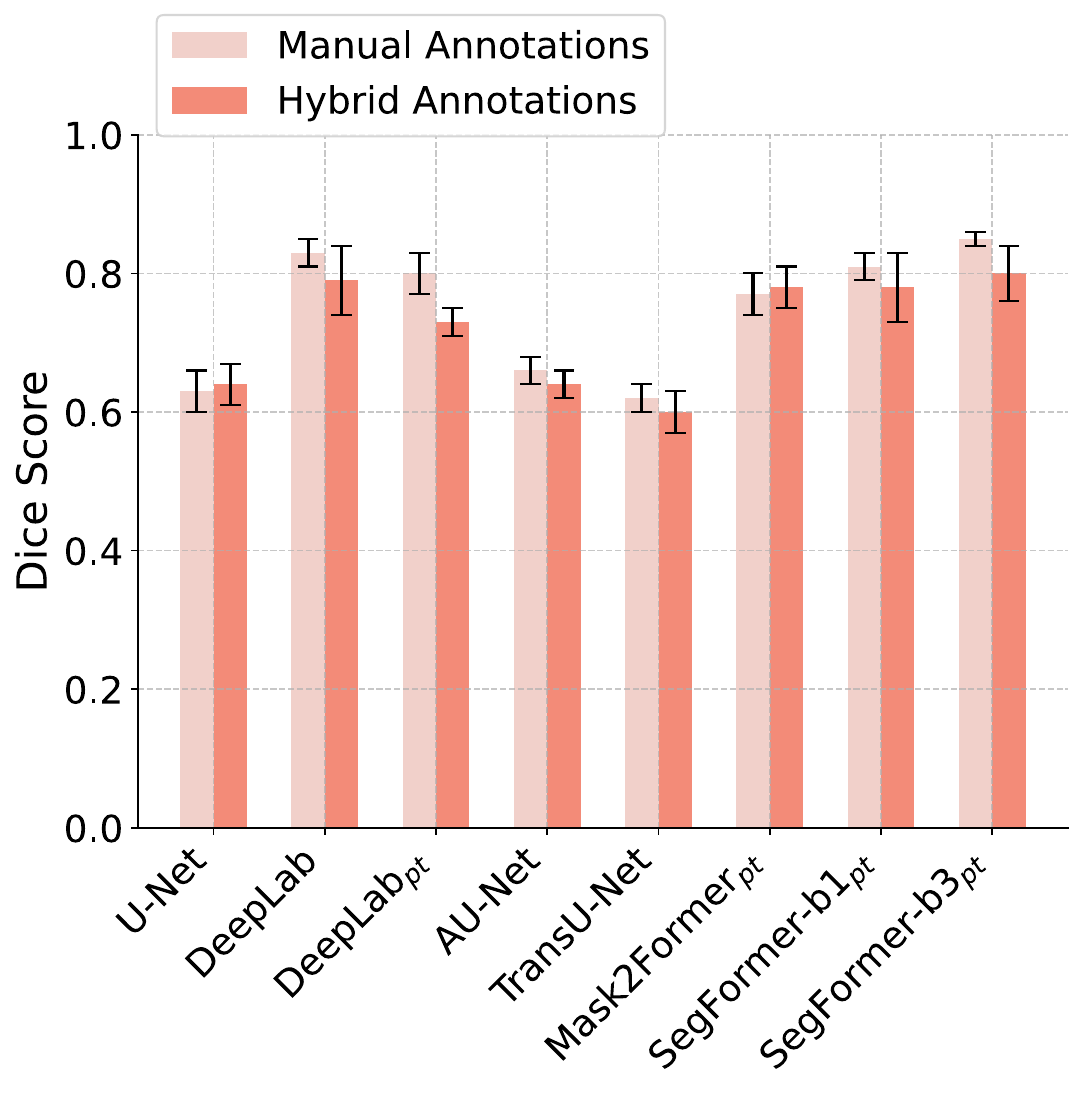}
        \caption{Test$_{Bloody}$}
        \label{fig:dice_score_manual_mix_t2}
    \end{subfigure}
    \hfill
    \begin{subfigure}[t]{0.32\textwidth}
        \centering
        \includegraphics[width=\textwidth]{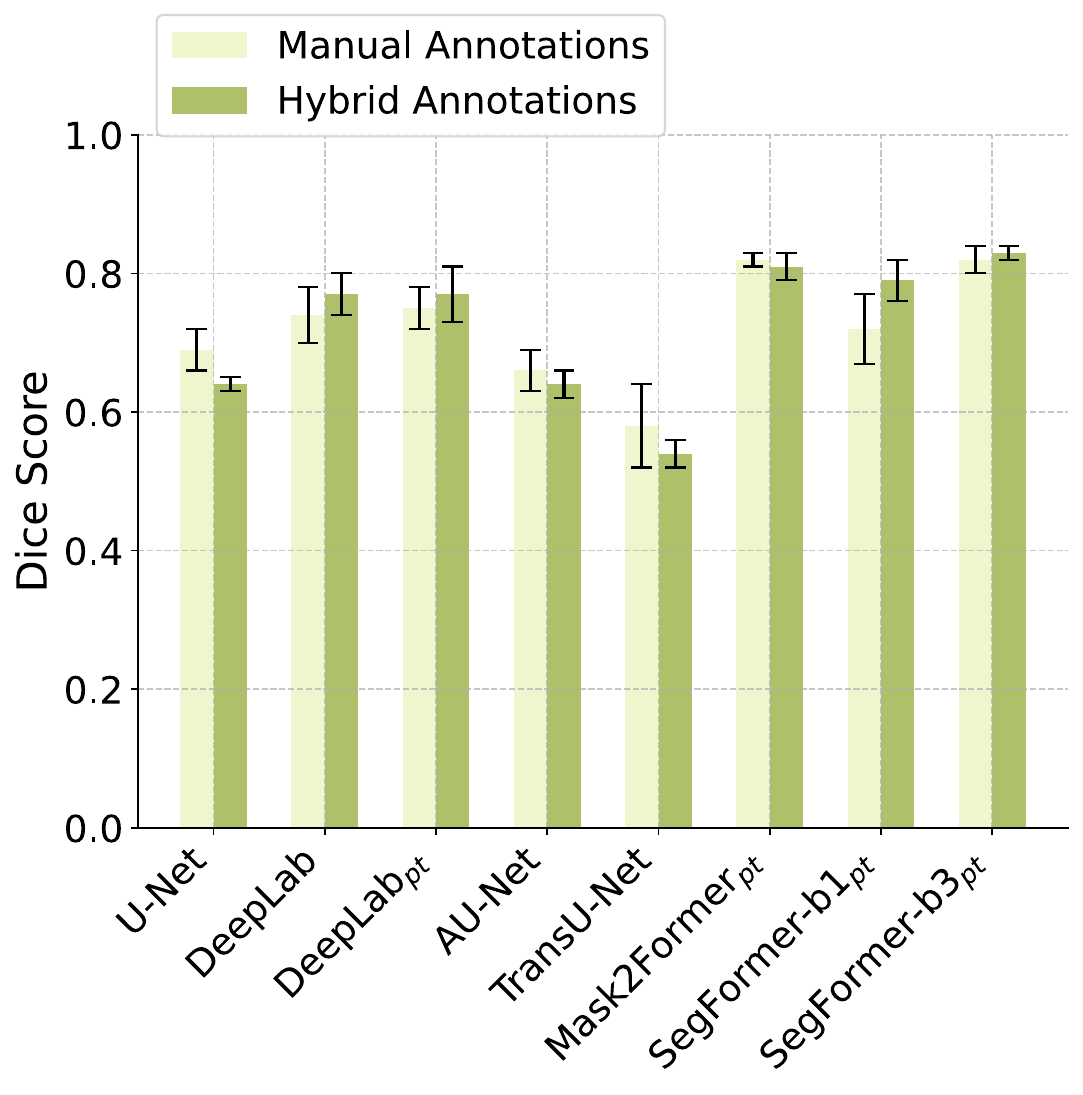}
        \caption{Test$_{Sim}$}
        \label{fig:dice_score_manual_mix_t3}
    \end{subfigure}
\caption{Dice scores on test splits for model training with manually and hybridly annotated samples. Performance gains are achieved on Test$_{Generic}$ for all cases except SegFormer-b3$_{pt}$. U-Net and AU-Net benefit from the feature knowledge of auto-tracked annotations in Test$_{Generic}$. Models with superior performance on manual annotations achieve marginal fluctuations, possibly due to performance saturation.}
\label{fig:dice_score_manual_mix}
\end{figure*}

\begin{figure*}[!ht]
    \centering
    \begin{subfigure}[t]{0.32\textwidth}
        \centering
        \includegraphics[width=\linewidth]{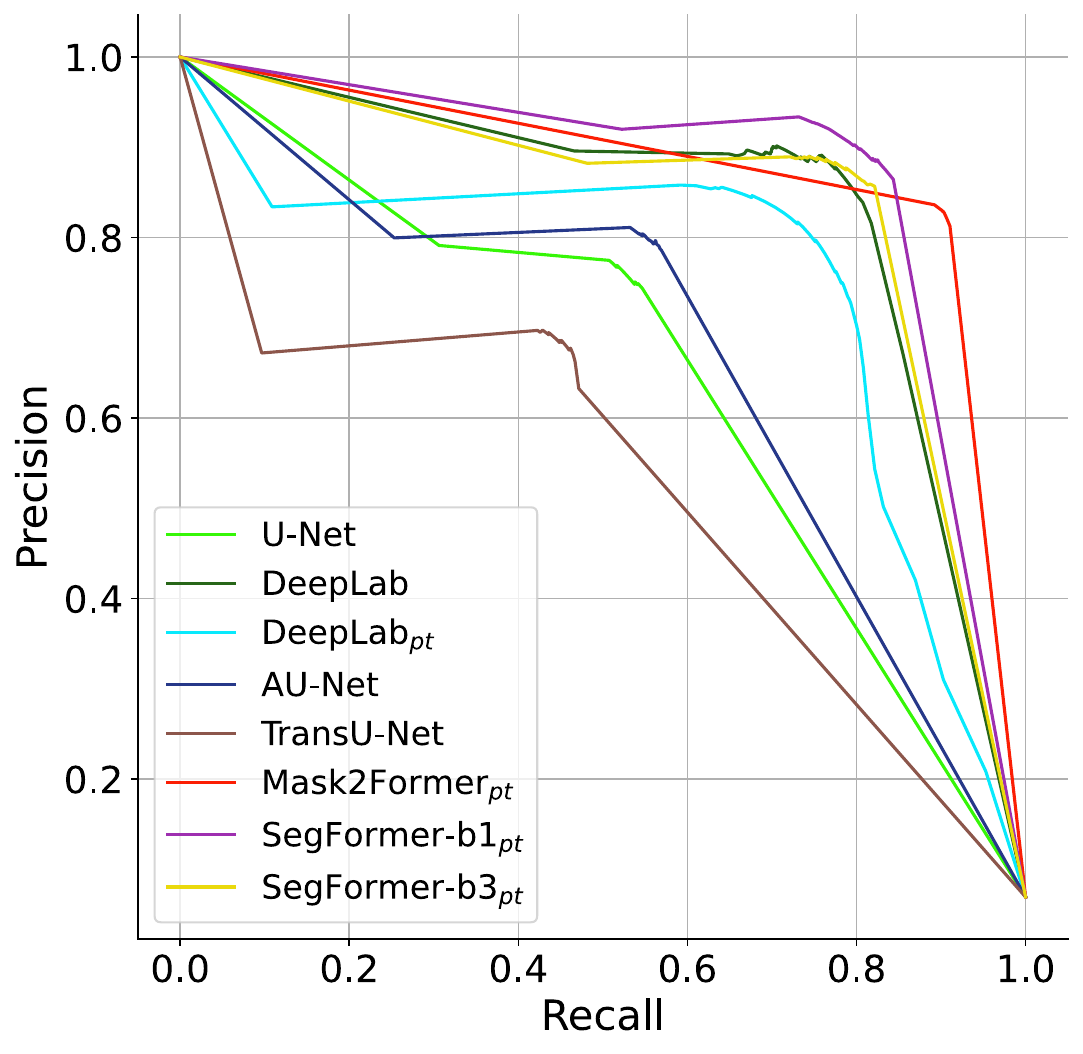}
        \caption{Test$_{Generic}$}
        \label{fig:pr_curve_mix_t1}
    \end{subfigure}
    \begin{subfigure}[t]{0.32\textwidth}
        \centering
        \includegraphics[width=\linewidth]{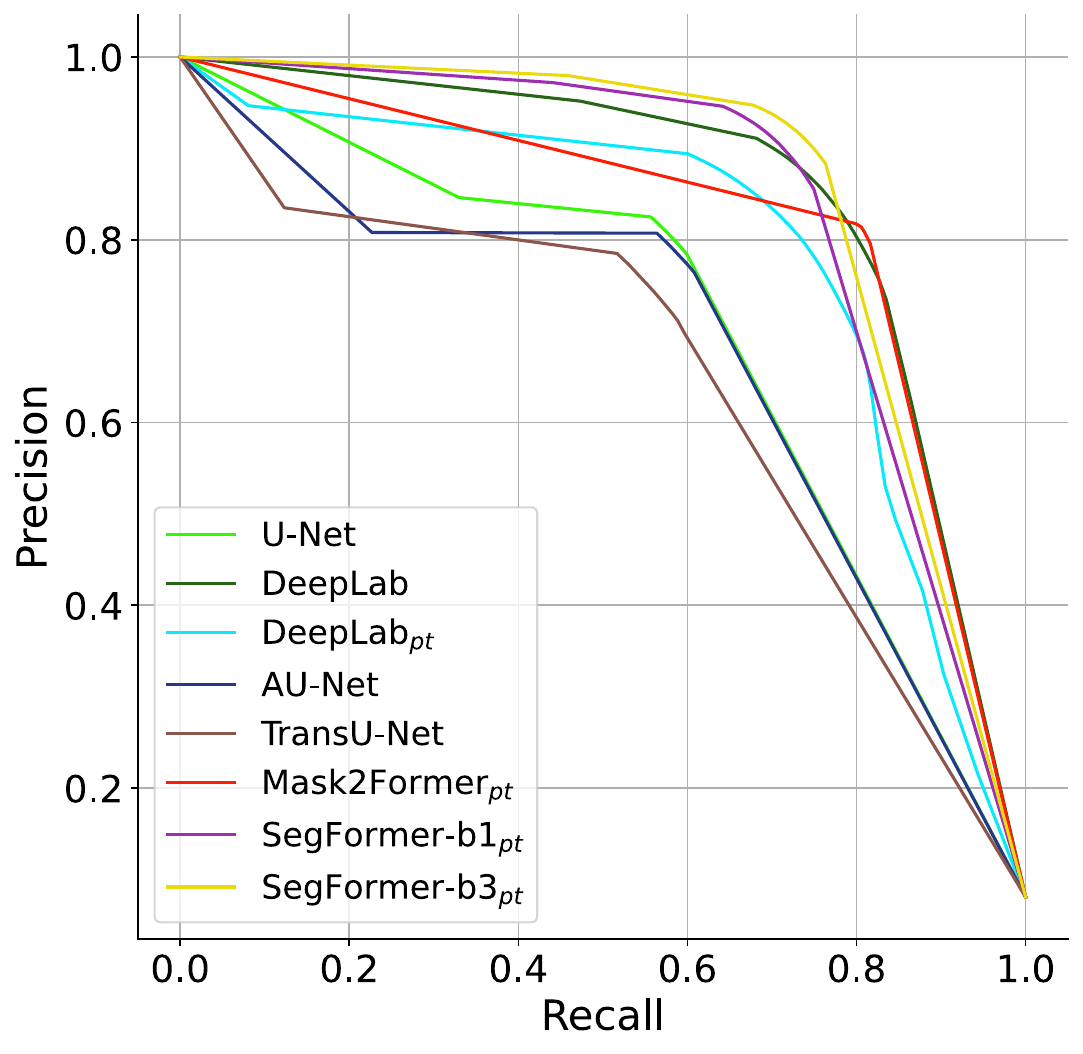}
        \caption{Test$_{Bloody}$}
        \label{fig:pr_curve_mix_t2}
    \end{subfigure}
    \begin{subfigure}[t]{0.32\textwidth}
        \centering
        \includegraphics[width=\linewidth]{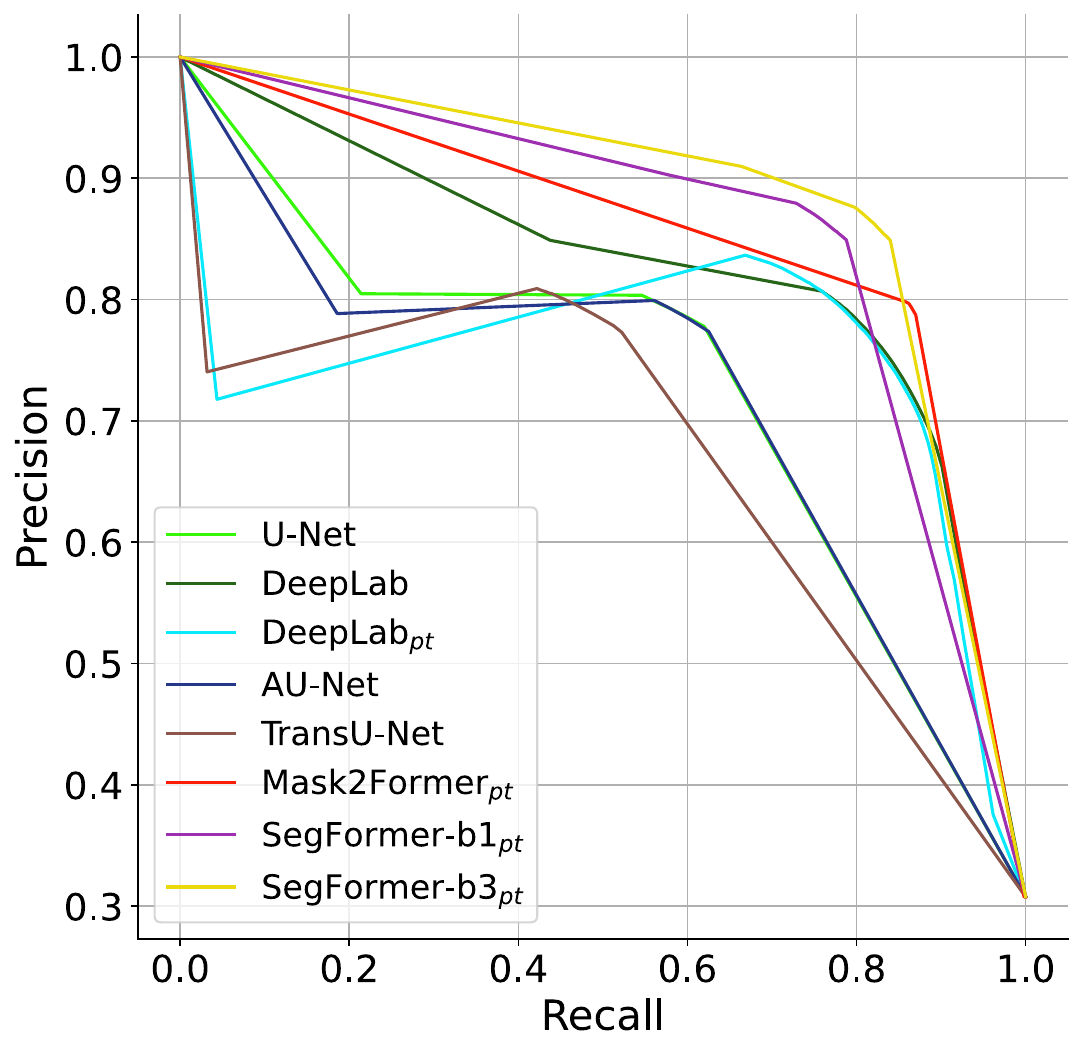}
        \caption{Test$_{Sim}$}
        \label{fig:pr_curve_mix_t3}
    \end{subfigure}
    \caption{Precision-Recall curves for models trained on hybrid annotations. Values represent the mean over 5 training runs per architecture. Fine-tuned SegFormer-b3$_{pt}$ achieves superior performance.}
    \label{fig:pr_curve_mix}
\end{figure*}

We initially investigate the segmentation performance of our considered architectures solely on the manually annotated samples. We present the Dice and IoU scores in Table~\ref{tab:baseline} and precision-recall (PR) curve plots in Figure~\ref{fig:pr_curve_manual}. First, the results reflect superior scores for all models on \mbox{Test$_{Bloody}$} compared to \mbox{Test$_{Generic}$}. This is a notably favorable outcome in contrast to the results reported in previous research on gauze detection, where the presence of blood makes detection challenging due to textural distortion and visual blending with anatomical tissues~\cite{ref_gauze_track0,ref_gauze_track,ref_gauze_track_1,ref_gauze_track5}. This highlights the significance of learning from real-world domain-specific data, as the widespread blood presence in the training data enabled effective learning of blood-saturated gauze features.

Second, the Dice and IoU scores on \mbox{Test$_{Sim}$} indicate robustness and effectiveness on out-of-distribution (OOD) data due to learned contextual knowledge. \mbox{Test$_{Sim}$} yields comparatively higher scores than \mbox{Test$_{Generic}$} for the majority of cases except SegFormer. Third, among pure CNN architectures including U-Net and DeepLab, the latter demonstrates significantly better scores, more prominently on in-domain test sets. Additionally, the effect of transfer learning on DeepLab reflects no major differences.

The fourth observation pertains to U-Net-based pure CNN or hybrid architectures, that is, \mbox{U-Net}, \mbox{AU-Net}, and \mbox{TransU-Net}, showing similar results, which can presumably be attributed to the underlying training framework and similarity in model capacities. This can be further examined in future work.

Lastly, the pretrained SegFormer models clearly outperform other architectures, with \mbox{SegFormer-b3$_{pt}$} exhibiting the best results on all test splits. Furthermore, the effect of model size can be observed for pretrained SegFormer, where the larger model \mbox{SegFormer-b3$_{pt}$} outperforms \mbox{SegFormer-b1$_{pt}$}. Of note, transformer architectures reflect their data-hungry nature, where training from scratch yields inadequate performance.

\subsection{Hybrid Annotations}
To investigate the effect of additional auto-tracked annotations, we train the models with samples having hybrid annotations, that is, a mix of manual and auto-tracked annotations. We present the variation in Dice score for manual and hybrid annotations in Figure~\ref{fig:dice_score_manual_mix} and Figure~\ref{fig:pr_curve_mix} respectively. Refer to Table~\ref{tab:results_mix} for Dice and IoU scores. Consistent with the prior results, \mbox{SegFormer-b3$_{pt}$} illustrates overall superior performance, highest on Test$_{Bloody}$ and \mbox{Test$_{Sim}$} sets. Figure~\ref{fig:segformer_all} presents segmentation masks of \mbox{SegFormer-b3$_{pt}$} on the test splits. In addition to the comparatively larger gauzes in Test$_{Sim}$, the model is able to segment small-sized gauzes in Test$_{Generic}$ and Test$_{Bloody}$.

Second, we observe a considerable improvement for all models (except \mbox{SegFormer-b3$_{pt}$}) on the Test$_{Generic}$ set. In particular, CNN-variants trained from scratch exhibit pronounced improvements of at least 7\% (14\% for AU-Net) in Dice score. As anticipated, this indicates that auto-tracked labels have incorporated novel contextual information leading to better generalization, despite their inferior annotation quality.

The third key insight is the decrease in performance on \mbox{Test$_{Bloody}$} for the majority of cases, with a noticeable drop in \mbox{SegFormer$_{pt}$} and DeepLab variants. However, the aggregate improvement on \mbox{Test$_{Generic}$} outweighs this performance decline. This suggests an increase of generic (i.e., clean or partially blood-stained) distribution in the training data, emphasizing the relevance of data-driven learning in facilitating clinical applicability. Furthermore, the performance degradation on the OOD test set \mbox{Test$_{Sim}$} and noticeable improvement on Test$_{Generic}$ for U-Net variants (U-Net, AU-Net, and TransU-Net) reflect that these architectures are highly sensitive to the training data distribution. Figure~\ref{fig:u-net} presents the gauze masks for U-Net trained with manual and hybrid annotations, reflecting improvement in clean gauze prediction masks.

\subsection{Segmentation Boundary Analysis}
Overlap-based metrics, Dice and IoU, primarily measure the overlap between prediction and target annotation masks. Prior works have reported a trade-off between blood presence and gauze detection due to the loss of textural information and blending with anatomical tissues~\cite{ref_gauze_track,ref_gauze_track0,ref_gauze_track_1,ref_gauze_track5}. Moreover, it is imperative to predict well-defined boundaries of gauze for various downstream applications. Therefore, to assess the quality of mask contours, we use normalized surface distance (NSD)~\cite{nikolov2021clinically,seidlitz2022robust} as recommended by the Metrics Reloaded platform~\cite{Maier_Hein_2024}. NSD measures the fraction of pixel boundary of the annotation mask lying within a distance threshold $\tau$ of the reference mask. Let $\mathcal{D}_{Y}$ represent the set of nearest-neighbor distances from the predicted mask boundary $\hat{Y}$ to the ground truth mask boundary $Y$, and $\mathcal{D}_{\hat{Y}}$ the distances calculated vice-versa. $\mathcal{D}_{Y}^{'}$ and $\mathcal{D}_{\hat{Y}}^{'}$ are the subsets containing distances within the threshold $\tau$. NSD can be computed as:

\begin{equation*}
     \operatorname {NSD} \left(Y, \hat{Y}\right) = \frac{\left|\mathcal{D}_{Y}^{'}\right| +
        \left| \mathcal{D}_{\hat{Y}}^{'} \right|}{\left|\mathcal{D}_{Y}\right| +
        \left|\mathcal{D}_{\hat{Y}}\right|} 
\end{equation*}
where
\begin{equation*}
    \mathcal{D}_{Y}^{'} = \{ d \in \mathcal{D}_{Y} \, | \, d \leq \tau \}
\end{equation*}
            
\begin{figure}[!tb]
    \centering
     \includegraphics[width=0.5\linewidth]{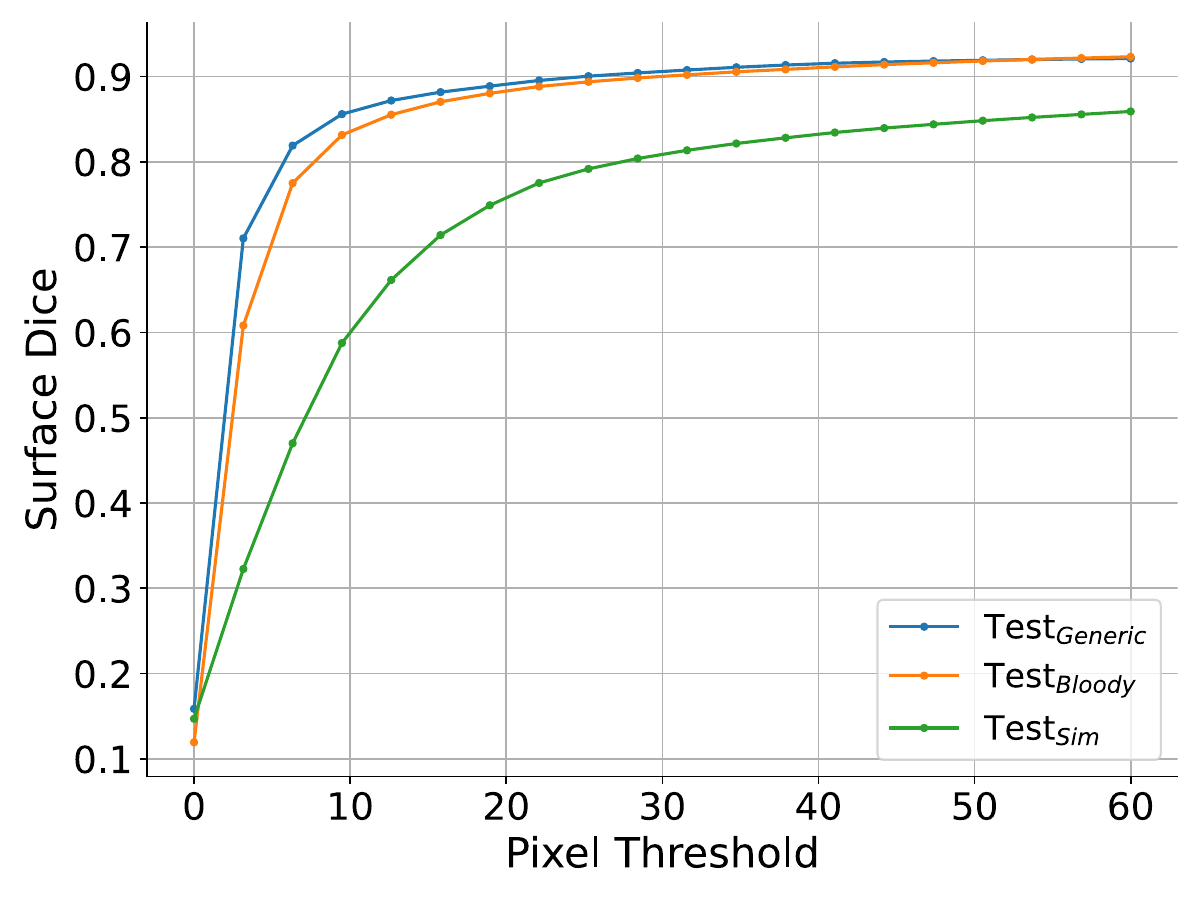}
    \caption{Surface Dice Scores of \mbox{SegFormer-b3$_{pt}$} on three test sets for different threshold values. The model shows better results for in-domain test sets than the OOD Test$_{Sim}$ set. The texture and spatial features of gauze in GauzeSeg$_{Real}$ are highly divergent from GauzeSeg$_{Sim}$, presenting challenges in inferring precise boundaries.}
    \label{fig:surface_dice} 
\end{figure}

Similar to the Dice score, NSD $\in [0,1]$ where 1 denotes a perfect match of segmentation boundaries. We evaluate NSD for the best-performing pretrained SegFormer-b3$_{pt}$ model with different distance thresholds (Figure~\ref{fig:surface_dice}) following the MONAI~\cite{cardoso2022monaiopensourceframeworkdeep} implementation, which evaluates to an undefined value if a given class is missing from both prediction and ground truth. As the gauze class is present in all samples of test sets Test$_{Generic}$ and Test$_{Bloody}$, we evaluate NSD on the full test sets. However, Test$_{Sim}$ contains 20 samples with no gauze masks; therefore, we remove no-gauze predictions for these no-gauze target masks.

\newcommand{\segresult}[3]{%
  \begin{subfigure}[t]{\columnwidth}
    \centering
    \sbox0{\includegraphics[width=0.87\columnwidth]{#1}}%
    \begin{minipage}[c][\ht0]{0.10\columnwidth}
      \raggedright
      \vspace{0.15\ht0}
      \scriptsize Input
      \vfill
      \scriptsize Target
      \vfill
      \scriptsize Prediction
      \vspace{0.2\ht0}
    \end{minipage}%
    \hspace{3pt}%
    \begin{minipage}[c]{0.87\columnwidth}
      \usebox0
    \end{minipage}
    \caption{#2}
    \label{#3}
  \end{subfigure}
}

\begin{figure*}[!p]
    \centering
    \segresult{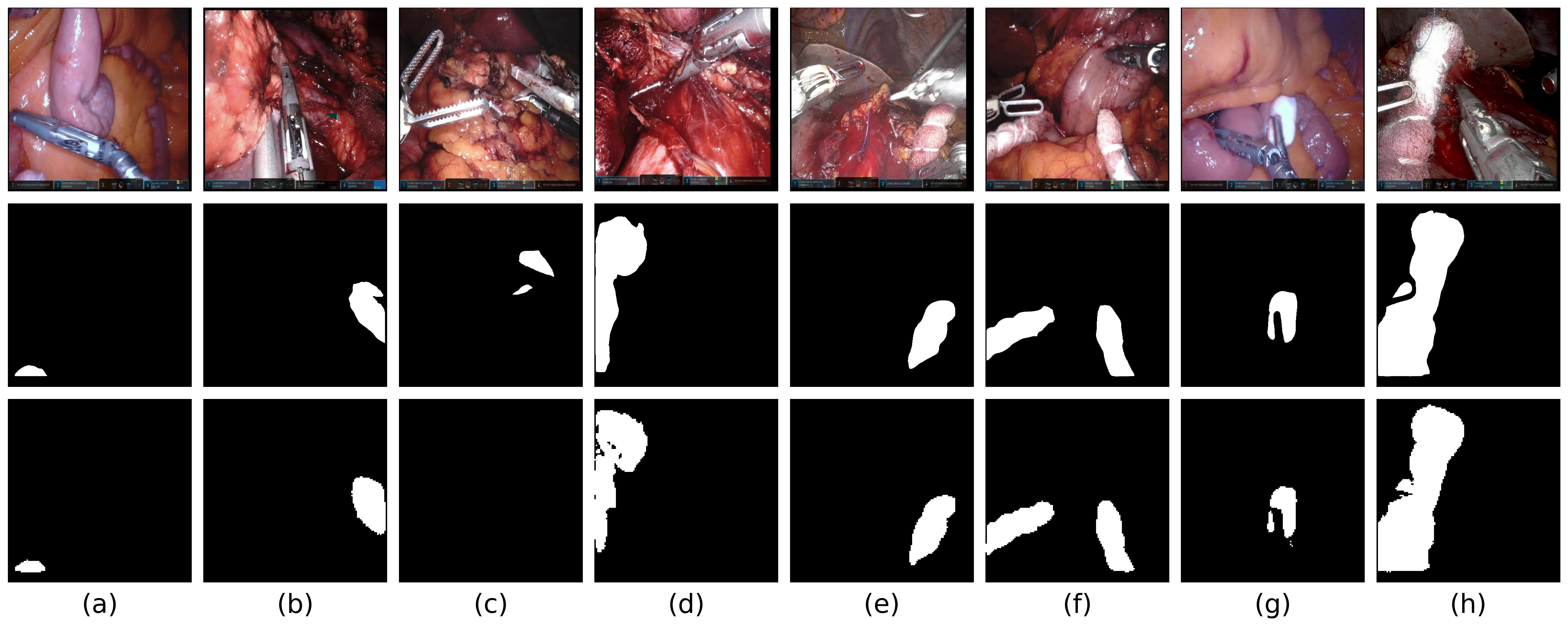}{Test$_{Generic}$}{fig:segformer_t1}
    \vspace{4pt}
    \segresult{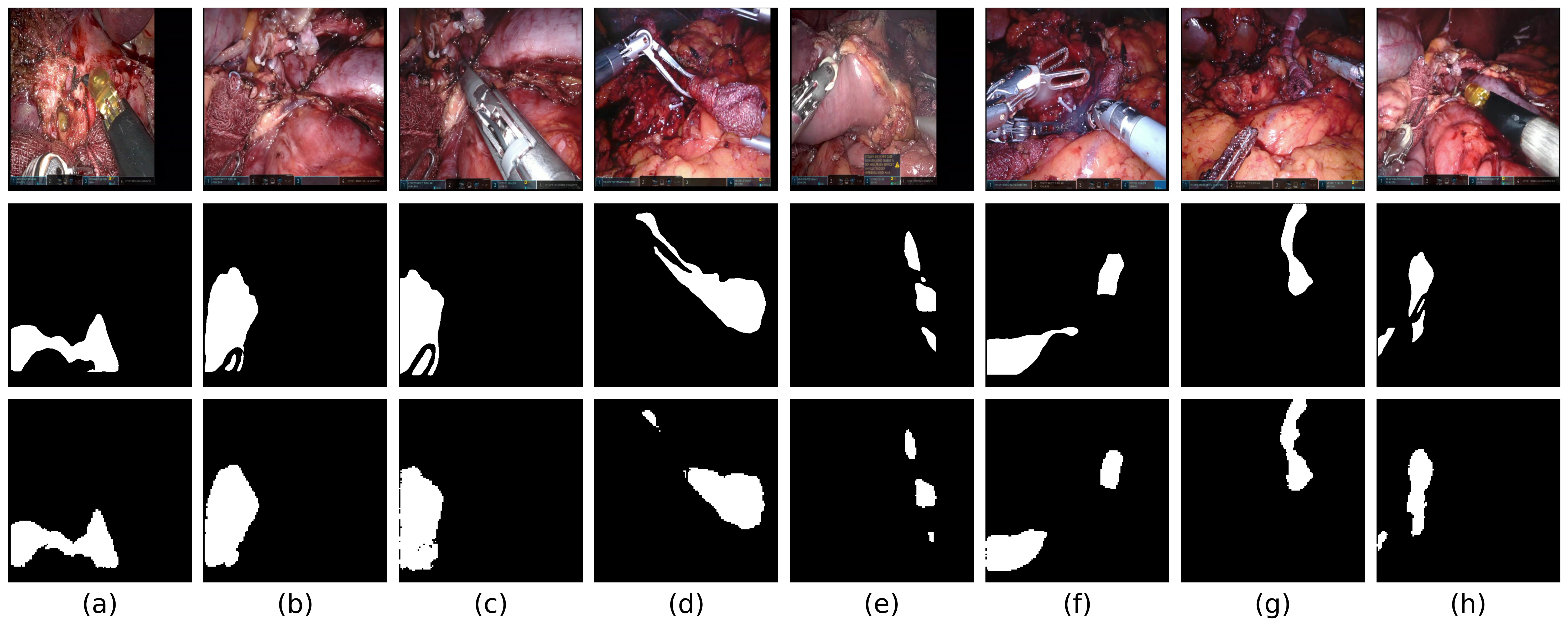}{Test$_{Bloody}$}{fig:segformer_t2}
    \vspace{4pt}
    \segresult{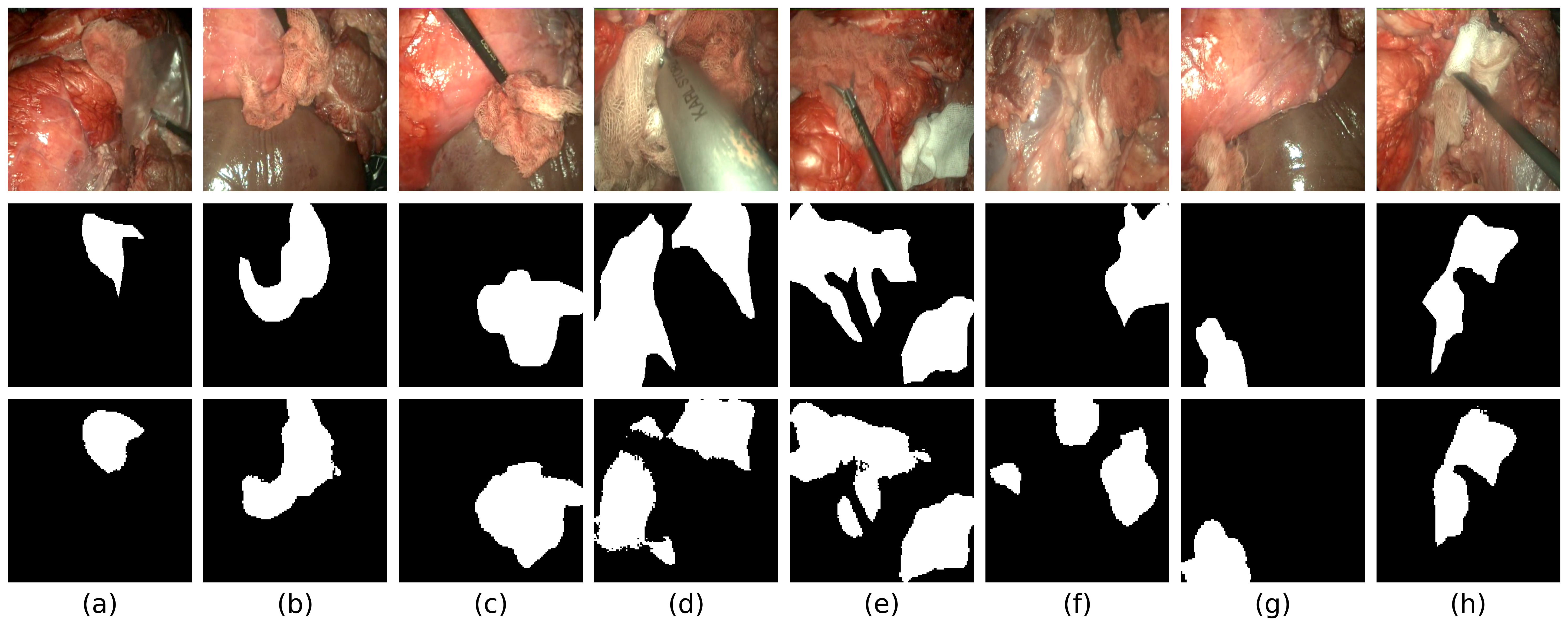}{Test$_{Sim}$}{fig:segformer_t3}

    \caption{Qualitative segmentation results of SegFormer-b3$_{pt}$ trained on hybrid annotations across the three test sets. Each subfigure shows the input image (top row), ground truth mask (middle row), and model prediction (bottom row). The model demonstrates effective gauze segmentation across diverse scenarios, including varying gauze sizes, blood presence, and out-of-distribution simulated data.}
    \label{fig:segformer_all}
\end{figure*}

Considering the plot, NSD scores are highest for Test$_{Generic}$, in contrast to the trend in Table~\ref{tab:baseline} where Dice scores are higher for Test$_{Bloody}$ compared to Test$_{Generic}$. This indicates that predicting correct gauze contours is more difficult in the presence of blood due to the loss of textural information. We attribute this in part to the use of dice loss for training, which is a region-based loss. We plan to incorporate boundary information in the loss function in future work.

The improvements in NSD values for Test$_{Generic}$ and Test$_{Bloody}$ are marginal after the threshold of 30 pixels, indicating that the detected mask boundaries lie in close vicinity of the true mask, and the NSD scores exceed 70\% for a threshold value of 10 pixels. Concerning Test$_{Sim}$, the model demonstrates the lowest scores, which can be possibly attributed to domain shift and larger mask perimeter. This serves as a direction for future investigation.

\subsection{Architectural Comparison}
For transformer models, learning through a large amount of data is more important for performance than model capacity, which can be leveraged through transfer learning. For CNN-based models, model size considerably influenced performance over pretrained features. DeepLab outperformed vanilla U-Net architectures but did not offer evident gains when utilizing pretrained weights. This can be attributed to the inherent learning mechanisms of CNN and transformer architectures: CNNs focus on learning from local context, performing well for smaller datasets by learning fine-grained details but not retaining pre-acquired knowledge in domain-shift settings. Transformers focus on global data features and therefore require additional data for learning localized features, but they show better retention of pretrained knowledge for transfer learning.

\section{Limitations}
While this work presents a first investigation into gauze segmentation on real surgical data, several limitations should be acknowledged. First, the GauzeSeg$_{Real}$ dataset originates from a single institution (University Hospital Bonn) and covers 16 robot-assisted laparoscopic surgeries. Although we ensure patient-level splits for evaluation, the generalizability of the findings to other hospitals, surgical teams, or different types of minimally invasive procedures remains to be validated through multi-center studies. Second, the current study addresses binary gauze segmentation only, treating all three gauze categories (conventional gauze, cigar gauze, and swab) as a single foreground class. This simplification does not capture the clinically relevant distinctions between gauze types, which may matter for downstream tasks such as gauze counting or usage tracking. Third, the auto-tracked annotations, while beneficial for augmenting training data, introduce noise in the form of imprecise object contours. We have not explicitly modeled or mitigated this label noise, for instance through noise-robust loss functions or curriculum learning strategies, which could further improve performance. Fourth, the hyperparameter configurations used in this study were determined through limited initial experiments due to computational constraints, and a more exhaustive search could yield improved results, particularly for architectures that showed sensitivity to training settings. Finally, we have not yet evaluated real-time inference latency, which is a critical consideration for clinical deployment in intraoperative settings where timely feedback is essential.

\section{Conclusion and Future Work}
In conclusion, we present the first investigation of surgical gauze segmentation on real-world data. Our training data reflect real operative conditions in robot-assisted laparoscopic surgeries, including extensive heterogeneity in structural, contextual, visual, and spatial aspects of gauze. This facilitated positive outcomes, particularly in overcoming the influence of high blood presence, and enabled robust performance on out-of-distribution simulated data.

Favorable performance and clinical applicability of deep learning models demand high-volume domain-specific data. However, high-quality data annotation in medical imaging involves specialized knowledge and extensive efforts. We followed a hybrid approach by using auto-tracked annotations, which incorporate in-domain knowledge but inferior quality annotations, and address the bottleneck of data scarcity. Our results present insights into the effectiveness of auto-tracked annotation in the surgical domain, which is an emerging research area~\cite{hakkinen2024medical}.

In addition to CNN models, we investigated transformer-based and hybrid architectures for gauze segmentation, which have not been explored even for gauze detection. Hence, our work presents a foundational contribution and provides insights into the potential of these architectures.

Future work will focus on in-depth investigation into dataset quality, comprehensive validation, inference latency, and model capacities for clinical applicability. As this research explores the potential of different architectures for gauze segmentation, we plan extensive hyperparameter optimizations of promising models such as DeepLab and transformers in follow-up work. Furthermore, we believe that additional state-of-the-art architectures and self-optimized frameworks such as nnU-Net~\cite{isensee2021nnu} can further enhance performance. Addressing the label noise introduced by auto-tracked annotations through noise-aware training strategies is another promising direction.
This research presents a preliminary examination of binary gauze segmentation. Our dataset contains three different gauze categories and the co-existence of multiple gauzes in a single frame. Therefore, future work can explore multi-class and instance segmentation for gauze counting and tracking, providing an advanced system for downstream applications. Multi-center validation will also be essential to establish the broader clinical applicability of the proposed approaches.
Concerning data dependence, we suggest self-supervised pre-training on surgical datasets such as Cholec80~\cite{endonet} and Endovis\footnote{\href{https://opencas.dkfz.de/endovis/datasetspublications/}{Endoscopic Vision Datasets}} for general feature learning, as well as data augmentation through generative models.

\section*{Acknowledgment}
\begin{itemize}
    \item This research has been funded by the Federal Ministry of Education and Research of Germany and the state of North-Rhine Westphalia as part of the Lamarr-Institute for Machine Learning and Artificial Intelligence, LAMARR22B.
    \item This research is funded by the Deutsche Forschungsgemeinschaft (DFG, German Research Foundation) under Germany's Excellence Strategy---EXC 2070-390732324-PhenoRob.
    \item The research leading to these results received funding from the Ministry of Economic Affairs, Industry, Climate Action and Energy of the State of North Rhine-Westphalia, Germany, through the ``Innovative Secure Medical Campus'' (ISMC) Project.
\end{itemize}

\section*{Conflicts of Interest}
\begin{itemize}
    \item J.~Arensmeyer and P.~Feodorovici declare that they hold minority interest in Medicalholodeck AG. J.~Arensmeyer and P.~Feodorovici received travel support from Medtronic Germany GmbH, Medicalholodeck AG, and Distalmotion SA.
    \item J.~Arensmeyer declares that he has received a speaker's honorarium from Medicalholodeck AG, Medtronic Germany GmbH, and Chiesi GmbH.
    \item P.~Feodorovici and J.~Arensmeyer received advisory fees from Richard Wolf GmbH.
\end{itemize}

\bibliographystyle{unsrtnat}
\bibliography{bibliography}

@inproceedings{schneider2022improving,
  title={Improving Intensive Care Chest X-Ray Classification by Transfer Learning and Automatic Label Generation.},
  author={Schneider, Helen and others},
  booktitle={ESANN},
  year={2022}
}

@inproceedings{schneider2023one,
  title={Is one label all you need? Single positive multi-label training in medical image analysis},
  author={{H. Schneider} and others},
  booktitle={2023 IEEE International Conference on Big Data (BigData)},
  pages={4963--4972},
  year={2023},
  organization={IEEE}
}

@inproceedings{schneider2023segmentation,
  title={Segmentation and analysis of lumbar spine mri scans for vertebral body measurements},
  author={Schneider,Helen and others},
  booktitle={The European Symposium on Artificial Neural Networks},
  year={2023}
}

@inproceedings{schneider2024informed,
  title={Informed Deep Abstaining Classifier: Investigating Noise-Robust Training for Diagnostic Decision Support Systems Systems},
  author={{H. Schneider} and others},
  booktitle={International Conference on Neural Information Processing},
  pages={402--416},
  year={2024},
  organization={Springer}
}

@inproceedings{tomar2025effective,
  title={Effective Disjoint Representational Learning for Anatomical Segmentation},
  author={Tomar, Priya and others},
  booktitle={Medical Imaging with Deep Learning},
  year={2025}
}

@article{hofman2024first,
  title={First-in-human real-time AI-assisted instrument deocclusion during augmented reality robotic surgery},
  author={Hofman, Jasper and others},
  journal={Healthcare Technology Letters},
  volume={11},
  number={2-3},
  pages={33--39},
  year={2024},
  publisher={Wiley Online Library}
}

@inproceedings{ref_gauze_diag1,
  title={Automated identification of retained surgical items in radiological images},
  author={Agam, Gady and Gan, Lin and Moric, Mario and Gluncic, Vicko},
  booktitle={Medical Imaging 2015: Pacs and Imaging Informatics: Next Generation and Innovations},
  volume={9418},
  pages={206--212},
  year={2015},
  organization={SPIE}
}

@article{ref_gauze_diag2,
  title={Minimization of occurrence of retained surgical items using machine learning and deep learning techniques: a review},
  author={Abo-Zahhad, Mohammed and El-Malek, Ahmed H Abd and Sayed, Mohammed S and Gitau, Susan Njeri},
  journal={BioData Mining},
  volume={17},
  number={1},
  pages={17},
  year={2024},
  publisher={Springer}
}

@inproceedings{ref_gauze_diag3,
  title={Surgical retained foreign object (RFO) prevention by computer aided detection (CAD)},
  author={Marentis, Theodore C and Hadjiiyski, Lubomir and Chaudhury, Amrita R and Rondon, Lucas and Chronis, Nikolaos and Chan, Heang-Ping},
  booktitle={Medical Imaging 2014: Computer-Aided Diagnosis},
  volume={9035},
  pages={590--595},
  year={2014},
  organization={SPIE}
}

@article{ref_gauze_det1,
title = "Bar-coding surgical sponges to improve safety: A randomized controlled trial",
author={Greenberg, Caprice C and others},
year = "2008",
month = apr,
doi = "10.1097/SLA.0b013e3181656cd5",
language = "English (US)",
volume = "247",
pages = "612--616",
journal = "Annals of surgery",
issn = "0003-4932",
publisher = "Wolters Kluwer Health",
number = "4",
}

@article{yoon2024automated,
  title={Automated deep learning model for estimating intraoperative blood loss using gauze images},
  author={Yoon, Dan and others},
  journal={Scientific Reports},
  volume={14},
  number={1},
  pages={2597},
  year={2024},
  publisher={Nature Publishing Group UK London}
}

@article{ref_gauze_det2,
  title={Designing a safer process to prevent retained surgical sponges: a healthcare failure mode and effect analysis},
  author={Steelman, Victoria M and Cullen, Joseph J},
  journal={AORN journal},
  volume={94},
  number={2},
  pages={132--141},
  year={2011},
  publisher={Elsevier}
}

@article{ref_gauze_det3,
  title={Radiofrequency-based identification medical device: an evaluable solution for surgical sponge retrieval?},
  author={Lazzaro, Alessandra and others},
  journal={Surgical Innovation},
  volume={24},
  number={3},
  pages={268--275},
  year={2017},
  publisher={SAGE Publications Sage CA: Los Angeles, CA}
}

@article{shyam2022gauze,
  title={Gauze for concern: A Case Report and systematic review of delayed presentation of paraspinal textiloma},
  author={Shyam, Karthik and Thippeswamy, Pushpa Bhari and Shetty, Ajoy Prasad and Algeri, Raksha and Rajasekaran, Shanmuganathan},
  journal={Journal of Clinical Orthopaedics and Trauma},
  volume={32},
  pages={101967},
  year={2022},
  publisher={Elsevier}
}

@article{rajkovic2010unusual,
  title={An unusual cause of chronic lumbar back pain: retained surgical gauze discovered after 40 years},
  author={Rajkovi{\'c}, Zoran and Altarac, Silvio and Pape{\v{s}}, Dino},
  journal={Pain medicine},
  volume={11},
  number={12},
  pages={1777--1779},
  year={2010},
  publisher={Blackwell Publishing Inc Malden, USA}
}

@article{pinto2023artificial,
  title={How artificial intelligence is shaping medical imaging technology: a survey of innovations and applications},
  author={Pinto-Coelho, Lu{\'\i}s},
  journal={Bioengineering},
  volume={10},
  number={12},
  pages={1435},
  year={2023},
  publisher={MDPI}
}

@article{pragadeeswari2020optical,
  title={Optical flow detection and tracking of gauzes and cancer cells},
  author={Pragadeeswari, C Karthika and Yamuna, G and Beham, G Yasmin},
  journal={Materials Today: Proceedings},
  volume={33},
  pages={3558--3563},
  year={2020},
  publisher={Elsevier}
}

@article{hariharan2013retained,
  title={Retained surgical sponges, needles and instruments},
  author={Hariharan, D and Lobo, DN},
  journal={The Annals of the Royal College of Surgeons of England},
  volume={95},
  number={2},
  pages={87--92},
  year={2013},
  publisher={Royal College of Surgeons}
}

@article{ref_LBP,
  title={Multiresolution gray-scale and rotation invariant texture classification with local binary patterns},
  author={Ojala, Timo and Pietikainen, Matti and Maenpaa, Topi},
  journal={IEEE Transactions on pattern analysis and machine intelligence},
  volume={24},
  number={7},
  pages={971--987},
  year={2002},
  publisher={IEEE}
}

@inproceedings{ref_gauze_track0,
  title={Automatic detection of surgical gauzes using Computer Vision},
  author={Garc{\'\i}a-Mart{\'\i}nez, {\'A}lvaro and Juan, Carlos G and Garcia, Nicolas M and Sabater-Navarro, Jos{\'e} Mar{\'\i}a},
  booktitle={2015 23rd Mediterranean Conference on Control and Automation (MED)},
  pages={747--751},
  year={2015},
  organization={IEEE}
}

@article{ref_gauze_track,
  title={Automatic gauze tracking in laparoscopic surgery using image texture analysis},
  author={de la Fuente L{\'o}pez, Eusebio and Garc{\'\i}a, {\'A}lvaro Mu{\~n}oz and Del Blanco, Lidia Santos and Marinero, Juan Carlos Fraile and Turiel, Javier P{\'e}rez},
  journal={Computer methods and programs in biomedicine},
  volume={190},
  pages={105378},
  year={2020},
  publisher={Elsevier}
}

@article{ref_gauze_track_1,
  title={Gauze detection and segmentation in minimally invasive surgery video using convolutional neural networks},
  author={S{\'a}nchez-Brizuela, Guillermo and others},
  journal={Sensors},
  volume={22},
  number={14},
  pages={5180},
  year={2022},
  publisher={MDPI}
}

@article{ref_gauze_track5,
  title={Intraoperative detection of surgical gauze using deep convolutional neural network},
  author={Lai, Shuo-Lun and Chen, Chi-Sheng and Lin, Been-Ren and Chang, Ruey-Feng},
  journal={Annals of Biomedical Engineering},
  volume={51},
  number={2},
  pages={352--362},
  year={2023},
  publisher={Springer}
}

@misc{gauze_seg_dataset,
    title={Dataset: Gauze detection and segmentation in minimally invasive surgery video using convolutional neural networks},
    author={S{\'a}nchez-Brizuela, Guillermo and de la Fuente-L{\'o}pez},
    url = { https://zenodo.org/records/6637871 },
    publisher = { Supervisely },
    year = { 2022 },
    doi = {10.5281/zenodo.6637871},
    month = { Jul },
    note = { visited on 2025-04-28 }
}

@inproceedings{redmon2016you,
  title={You only look once: Unified, real-time object detection},
  author={Redmon, J},
  booktitle={Proceedings of the IEEE conference on computer vision and pattern recognition},
  year={2016}
}

@article{rueckert2024methods,
  title={Methods and datasets for segmentation of minimally invasive surgical instruments in endoscopic images and videos: A review of the state of the art},
  author={Rueckert, Tobias and Rueckert, Daniel and Palm, Christoph},
  journal={Computers in Biology and Medicine},
  volume={169},
  pages={107929},
  year={2024},
  publisher={Elsevier}
}

@article{kamtam2025deep,
  title={Deep learning approaches to surgical video segmentation and object detection: A Scoping Review},
  author={Kamtam, Devanish N and others},
  journal={arXiv preprint arXiv:2502.16459},
  year={2025}
}

@inproceedings{ronneberger2015u,
  title={U-net: Convolutional networks for biomedical image segmentation},
  author={Ronneberger, Olaf and Fischer, Philipp and Brox, Thomas},
  booktitle={Medical image computing and computer-assisted intervention--MICCAI 2015: 18th international conference, Munich, Germany, October 5-9, 2015, proceedings, part III 18},
  pages={234--241},
  year={2015},
  organization={Springer}
}

@article{oktay2018attention,
  title={Attention u-net: Learning where to look for the pancreas},
  author={Oktay, Ozan and others},
  journal={arXiv preprint arXiv:1804.03999},
  year={2018}
}

@article{xie2021segformer,
  title={SegFormer: Simple and efficient design for semantic segmentation with transformers},
  author={Xie, Enze and Wang, Wenhai and Yu, Zhiding and Anandkumar, Anima and Alvarez, Jose M and Luo, Ping},
  journal={Advances in neural information processing systems},
  volume={34},
  pages={12077--12090},
  year={2021}
}

@inproceedings{cheng2022masked,
  title={Masked-attention mask transformer for universal image segmentation},
  author={Cheng, Bowen and Misra, Ishan and Schwing, Alexander G and Kirillov, Alexander and Girdhar, Rohit},
  booktitle={Proceedings of the IEEE/CVF conference on computer vision and pattern recognition},
  pages={1290--1299},
  year={2022}
}

@article{chen2021transunet,
  title={Transunet: Transformers make strong encoders for medical image segmentation},
  author={Chen, Jieneng and Lu, Yongyi and Yu, Qihang and Luo, Xiangde and Adeli, Ehsan and Wang, Yan and Lu, Le and Yuille, Alan L and Zhou, Yuyin},
  journal={arXiv preprint arXiv:2102.04306},
  year={2021}
}

@article{chen2017rethinking,
  title={Rethinking atrous convolution for semantic image segmentation. arXiv},
  author={Chen, Liang-Chieh and Papandreou, George and Schroff, Florian and Adam, Hartwig},
  journal={arXiv preprint arXiv:1706.05587},
  volume={5},
  year={2017}
}

@article{isensee2021nnu,
  title={nnU-Net: a self-configuring method for deep learning-based biomedical image segmentation},
  author={Isensee, Fabian and Jaeger, Paul F and Kohl, Simon AA and Petersen, Jens and Maier-Hein, Klaus H},
  journal={Nature methods},
  volume={18},
  number={2},
  pages={203--211},
  year={2021},
  publisher={Nature Publishing Group}
}

@inproceedings{isensee2024nnu,
  title={nnu-net revisited: A call for rigorous validation in 3d medical image segmentation},
  author={Isensee, Fabian and others},
  booktitle={International Conference on Medical Image Computing and Computer-Assisted Intervention},
  pages={488--498},
  year={2024},
  organization={Springer}
}

@article{endonet,
  author = {Twinanda, Andru P. and Shehata, Sherif and Mutter, Didier and Marescaux, Jacques and De Mathelin, Michel and Padoy, Nicolas},
  title = {EndoNet: A Deep Architecture for Recognition Tasks on Laparoscopic Videos},
  journal = {IEEE Transactions on Medical Imaging},
  volume = {36},
  year = {2016},
  month = feb,
  doi = {10.1109/TMI.2016.2593957}
}

@article{tanzi2021real,
  title={Real-time deep learning semantic segmentation during intra-operative surgery for 3D augmented reality assistance},
  author={Tanzi, Leonardo and Piazzolla, Pietro and Porpiglia, Francesco and Vezzetti, Enrico},
  journal={International Journal of Computer Assisted Radiology and Surgery},
  volume={16},
  number={9},
  pages={1435--1445},
  year={2021},
  publisher={Springer}
}

@article{Maier_Hein_2024,
   title={Metrics reloaded: recommendations for image analysis validation},
   volume={21},
   ISSN={1548-7105},
   DOI={10.1038/s41592-023-02151-z},
   number={2},
   journal={Nature Methods},
   publisher={Springer Science and Business Media LLC},
   author={Maier-Hein and others},
   year={2024},
   month=feb, pages={195–212} }

@article{nikolov2021clinically,
  title={Clinically applicable segmentation of head and neck anatomy for radiotherapy: deep learning algorithm development and validation study},
  author={Nikolov, Stanislav and others},
  journal={Journal of medical Internet research},
  volume={23},
  number={7},
  pages={e26151},
  year={2021},
  publisher={JMIR Publications Inc., Toronto, Canada}
}

@article{seidlitz2022robust,
  title={Robust deep learning-based semantic organ segmentation in hyperspectral images},
  author={Seidlitz, Silvia and Sellner, Jan and Odenthal, Jan and {\"O}zdemir, Berkin and Studier-Fischer, Alexander and Kn{\"o}dler, Samuel and Ayala, Leonardo and Adler, Tim J and Kenngott, Hannes G and Tizabi, Minu and others},
  journal={Medical Image Analysis},
  volume={80},
  pages={102488},
  year={2022},
  publisher={Elsevier}
}

@article{hakkinen2024medical,
  title={Medical Image Segmentation with SAM-generated Annotations},
  author={H{\"a}kkinen, Iira and Melekhov, Iaroslav and Englesson, Erik and Azizpour, Hossein and Kannala, Juho},
  journal={arXiv preprint arXiv:2409.20253},
  year={2024}
}

@misc{cardoso2022monaiopensourceframeworkdeep,
      title={MONAI: An open-source framework for deep learning in healthcare}, 
      author={Cardoso, M. Jorge and others},
      year={2022},
      eprint={2211.02701},
      archivePrefix={arXiv},
      primaryClass={cs.LG}, 
}
\appendix
\begin{table}[!h]
\caption{Results on hybrid annotations for different architectures. The abbreviation $pt$ indicates models initialized with pre-trained weights. \textbf{Bold} indicates the highest scores in the respective test sets. Comparing with Table~\ref{tab:baseline}, \textcolor{blue}{+} and \textcolor{blue}{++} indicate an increase of at least 5\% and 14\% in Test$_{Generic}$ mean scores. \textcolor{red}{$-$} denotes at least 5\% decrease in mean Test$_{Bloody}$ scores. \textcolor{teal}{$-$}/\textcolor{teal}{+} implies at least 5\% decrease/increase in mean Test$_{Sim}$ scores.}
\label{tab:results_mix}
\begin{center}
\begin{tabular}{llll}
\textbf{Model}   & \textbf{Test Set} & \textbf{Dice} & \textbf{IoU} \\ \toprule
\rowcolor{gray!20}U-Net &Test$_{Generic}$ & 0.57 $\pm$ 0.04 \textcolor{blue}{++} & 0.48 $\pm$ 0.03 \textcolor{blue}{++} \\
\rowcolor{gray!20}     &Test$_{Bloody}$ & 0.64 $\pm$ 0.03 +  & 0.50 $\pm$ 0.03 + \\
\rowcolor{gray!20}     &Test$_{Sim}$  & 0.64 $\pm$ 0.01 \textcolor{teal}{$-$} & 0.51 $\pm$ 0.01 \textcolor{teal}{$-$} \\ \midrule
\rowcolor{gray!20}DeepLab &Test$_{Generic}$  & 0.77 $\pm$ 0.04 \textcolor{blue}{+}   & 0.69 $\pm$ 0.04 \textcolor{blue}{+} \\
\rowcolor{gray!20}      &Test$_{Bloody}$ & 0.79 $\pm$ 0.05  &0.67 $\pm$ 0.05 \\
\rowcolor{gray!20}      &Test$_{Sim}$ & 0.77 $\pm$ 0.03 + & 0.66 $\pm$ 0.03 +\\ \midrule
\rowcolor{gray!20}DeepLab$_{pt}$ &Test$_{Generic}$ & 0.74 $\pm$ 0.01 \textcolor{blue}{+}  & 0.66 $\pm$ 0.02 \textcolor{blue}{+}\\
\rowcolor{gray!20}      &Test$_{Bloody}$  & 0.73 $\pm$ 0.02 \textcolor{red}{$-$}  & 0.60 $\pm$ 0.02 \textcolor{red}{$-$}\\
\rowcolor{gray!20}      &Test$_{Sim}$  & 0.77 $\pm$ 0.04 + & 0.66 $\pm$ 0.02 + \\ \specialrule{1pt}{4pt}{4pt}
\rowcolor{brown!10}AU-Net &Test$_{Generic}$ & 0.60 $\pm$ 0.02 \textcolor{blue}{++}  & 0.51 $\pm$ 0.02 \textcolor{blue}{++} \\
\rowcolor{brown!10}      &Test$_{Bloody}$ & 0.64 $\pm$ 0.02  & 0.50 $\pm$ 0.02 \\
\rowcolor{brown!10}      &Test$_{Sim}$  & 0.64 $\pm$ 0.02  & 0.52 $\pm$ 0.02 \\ \midrule
\rowcolor{brown!10}TransU-Net &Test$_{Generic}$ & 0.47 $\pm$ 0.01 \textcolor{blue}{+} & 0.40 $\pm$ 0.01 \textcolor{blue}{+}\\
\rowcolor{brown!10}      &Test$_{Bloody}$ & 0.60 $\pm$ 0.03 & 0.46 $\pm$ 0.03 \\
\rowcolor{brown!10}      &Test$_{Sim}$   & 0.54 $\pm$ 0.02  & 0.42 $\pm$ 0.02 \\ \specialrule{1pt}{4pt}{4pt}
Mask2Former &Test$_{Generic}$ & 0.12 $\pm$ 0.00  & 0.07 $\pm$ 0.00 \\
     &Test$_{Bloody}$ & 0.15 $\pm$ 0.00  & 0.08 $\pm$ 0.00 \\
      &Test$_{Sim}$  & 0.45 $\pm$ 0.00  & 0.31 $\pm$ 0.00 \\ \midrule
\rowcolor{red!5}Mask2Former$_{pt}$ &Test$_{Generic}$ & \textbf{0.84 $\pm$ 0.02} \textcolor{blue}{+} & \textbf{0.74 $\pm$ 0.03} \textcolor{blue}{+}\\
\rowcolor{red!5}      &Test$_{Bloody}$ &  0.78 $\pm$ 0.03 + & 0.66 $\pm$ 0.03 + \\
\rowcolor{red!5}      &Test$_{Sim}$  & 0.81 $\pm$ 0.02  & 0.70 $\pm$ 0.03 \\ \midrule
SegFormer-b1 & Test$_{Generic}$ & 0.41 $\pm$ 0.03 \textcolor{blue}{+} & 0.33 $\pm$ 0.02 \textcolor{blue}{+}\\
      &Test$_{Bloody}$ & 0.17 $\pm$ 0.07  & 0.10 $\pm$ 0.04 \\
      &Test$_{Sim}$ & 0.1 $\pm$ 0.05  & 0.07 $\pm$ 0.04 \\ \midrule
\rowcolor{red!8}SegFormer-b1$_{pt}$ &Test$_{Generic}$ & 0.83 $\pm$ 0.04 \textcolor{blue}{+} & 0.74 $\pm$  0.04 \textcolor{blue}{+} \\
\rowcolor{red!8}      &Test$_{Bloody}$  & 0.78 $\pm$ 0.05  & 0.66 $\pm$ 0.05 \\
 \rowcolor{red!8}     &Test$_{Sim}$  & 0.79 $\pm$ 0.03 \textcolor{teal}{+} & 0.68 $\pm$ 0.03 \textcolor{teal}{+}\\ \midrule
\rowcolor{red!8}SegFormer-b3$_{pt}$ &Test$_{Generic}$ & 0.80 $\pm$ 0.02  & 0.73 $\pm$ 0.02 \\
\rowcolor{red!8}      &Test$_{Bloody}$  & \textbf{0.80 $\pm$ 0.04} \textcolor{red}{$-$} &  \textbf{0.68 $\pm$ 0.05} \textcolor{red}{$-$} \\
\rowcolor{red!8}      &Test$_{Sim}$  & \textbf{0.83 $\pm$ 0.01} + & \textbf{0.73 $\pm$ 0.01} + \\
\end{tabular}
\end{center}
\end{table}

\begin{figure*}[h]
 \centering
  \sbox0{\includegraphics[width=0.85\textwidth]{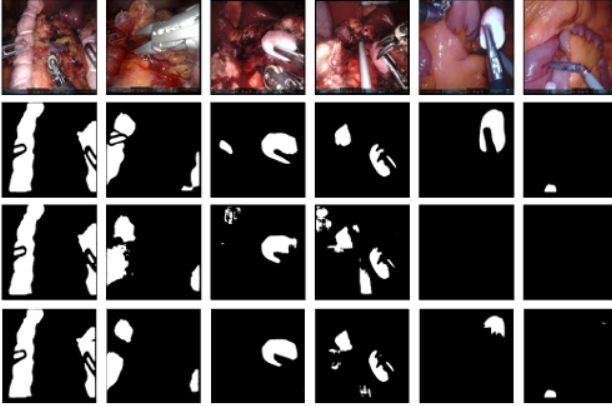}}%
  \begin{minipage}[c][\ht0]{0.08\textwidth}
    \raggedright
    \vspace{0.14\ht0}
    \scriptsize Input
    \vfill
    \scriptsize Target
    \vfill
    \scriptsize\raggedright Pred.\\ (Manual Anno.)
    \vfill
    \scriptsize\raggedright Pred.\\ (Hybrid Anno.)
    \vspace{0.06\ht0}
  \end{minipage}%
  \hspace{3pt}%
  \begin{minipage}[c]{0.85\textwidth}
    \usebox0
  \end{minipage}
  \caption{Improvement in segmentation results for the U-Net model on Test$_{Generic}$ set after incorporating auto-tracked annotations in the training data. The four rows show input images, ground truth masks, predictions from the model trained on manual annotations, and predictions from the model trained on hybrid annotations. The incorporation of auto-tracked annotations has resulted in improvements of mask quality for clean and partially stained gauze samples, with better object contours and a reduction in false positives.}
  \label{fig:u-net}
 \end{figure*}

\begin{figure}[!ht]
    \centering
    \includegraphics[width=\columnwidth]{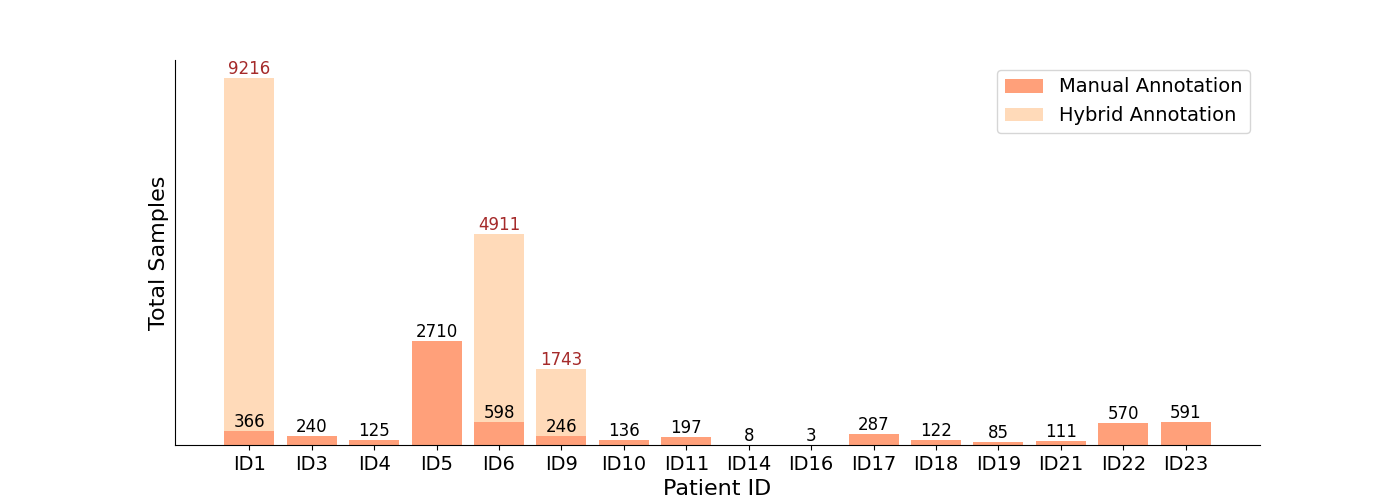}
    \caption{Distribution of manual and hybrid annotation samples per patient ID in the GauzeSeg$_{Real}$ dataset. Hybrid annotations substantially augment the training data, particularly for videos with longer gauze presence.}
    \label{fig:cls_distribution}
\end{figure}

\end{document}